\documentclass[final]{cvpr}

\usepackage{times}
\usepackage{epsfig}
\usepackage{graphicx}
\usepackage{amsmath}
\usepackage{amssymb}

\usepackage{multirow}
\usepackage{soul}
\usepackage{pifont} % http://ctan.org/pkg/pifont
\usepackage[T1]{fontenc}
\usepackage{mathabx,graphicx}
\usepackage{makecell}
\usepackage[dvipsnames]{xcolor}

\def\CircleArrowright{\ensuremath{%
  \rotatebox[origin=c]{310}{$\circlearrowright$}}}
\DeclareRobustCommand{\star}{%
  \begingroup\normalfont
  \includegraphics[height=\fontcharht\font`\B]{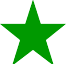}%
  \endgroup
}
\newcommand{\xmark}{\ding{55}} %
\newcommand\Tstrut{\rule{0pt}{2.3ex}}         % = `top' strut
\newcommand{\vlnbert}{VLN$\protect\CircleArrowright$BERT}
\newcommand{\high}[1]{{\textbf{\color{blue}#1}}}

\usepackage[pagebackref=true,breaklinks=true,colorlinks,bookmarks=false]{hyperref}

\def\cvprPaperID{784} % *** Enter the CVPR Paper ID here
\def\confYear{CVPR 2021}
\begin{document}

%%%%%%%%% TITLE
\title{\vlnbert: A Recurrent Vision-and-Language BERT for Navigation}

\author{
Yicong Hong$^1$ \quad Qi Wu$^2$ \quad Yuankai Qi$^2$ \quad Cristian Rodriguez-Opazo$^{1,2}$ \quad Stephen Gould$^1$\\
$^1$The Australian National University \quad $^2$The University of Adelaide\\
Australian Centre for Robotic Vision\\
{\tt\small \{yicong.hong, cristian.rodriguez, stephen.gould\}@anu.edu.au} \\
{\tt\small qi.wu01@adelaide.edu.au, qykshr@gmail.com} \\ \\
{\tt\small Project URL: \href{https://github.com/YicongHong/Recurrent-VLN-BERT}{https://github.com/YicongHong/Recurrent-VLN-BERT}}
}

\maketitle

% \pagestyle{empty}
% \thispagestyle{empty}

%%%%%%%%% ABSTRACT
\begin{abstract}
    % \SG{}
    Accuracy of many visiolinguistic tasks has benefited significantly from the application of vision-and-language (V\&L) BERT. However, its application for the task of vision-and-language navigation (VLN) remains limited. One reason for this is the difficulty adapting the BERT architecture to the partially observable Markov decision process present in VLN, requiring history-dependent attention and decision making. In this paper we propose a recurrent BERT model that is time-aware for use in VLN. Specifically, we equip the BERT model with a recurrent function that maintains cross-modal state information for the agent. Through extensive experiments on R2R and REVERIE we demonstrate that our model can replace more complex encoder-decoder models to achieve state-of-the-art results. Moreover, our approach can be generalised to other transformer-based architectures, supports pre-training, and is capable of solving navigation and referring expression tasks simultaneously.
\end{abstract}

%%%%%%%%% BODY TEXT

\section{Introduction}
\label{sec:introduction}

Asking a robot to navigate in complex environments following human instructions has been a long-term goal in AI research. Recently, a great variety of vision-and-language navigation (VLN) setups \cite{anderson2018vision, qi2020reverie, thomason2020vision} have been introduced for relevant studies and a large number of works explore different methods to leverage visual and language clues to assist navigation. For example, in the popular R2R navigation task \cite{anderson2018vision}, enhancing the learning of visual-textual correspondence is essential for the agent to correctly interpret the instruction and perceive the environment.

On the other hand, recent work on vision-and-language pre-training has achieved significant improvement over a wide range of visiolinguistic problems. Instead of designing complex and monolithic models for different tasks, those methods pre-train a multi-layer Transformer \cite{vaswani2017attention} on a large number of image-text pairs to learn generic cross-modal representations \cite{chen2020uniter,li2020unicoder,li2019visualbert,li2020oscar, lu2019vilbert,su2019vl,tan2019lxmert}, known as V\&L BERT (Bidirectional Encoder Representations from Transformers \cite{devlin2019bert}). Such advances have inspired us to employ V\&L BERT for VLN, replacing the complicated modules for modelling cross-modal relationships and allowing the learning of navigation to adequately benefit from the pre-trained visual-textual knowledge. Unlike recent works on VLN, which apply a pre-trained V\&L BERT only for encoding language \cite{hao2020towards,li2019robust} or for measuring the instruction-path compatibility \cite{majumdar2020improving}, we propose to use existing V\&L BERT models themselves for learning to navigate.

\begin{figure}[t]
  \centering
  \includegraphics[width=\columnwidth]{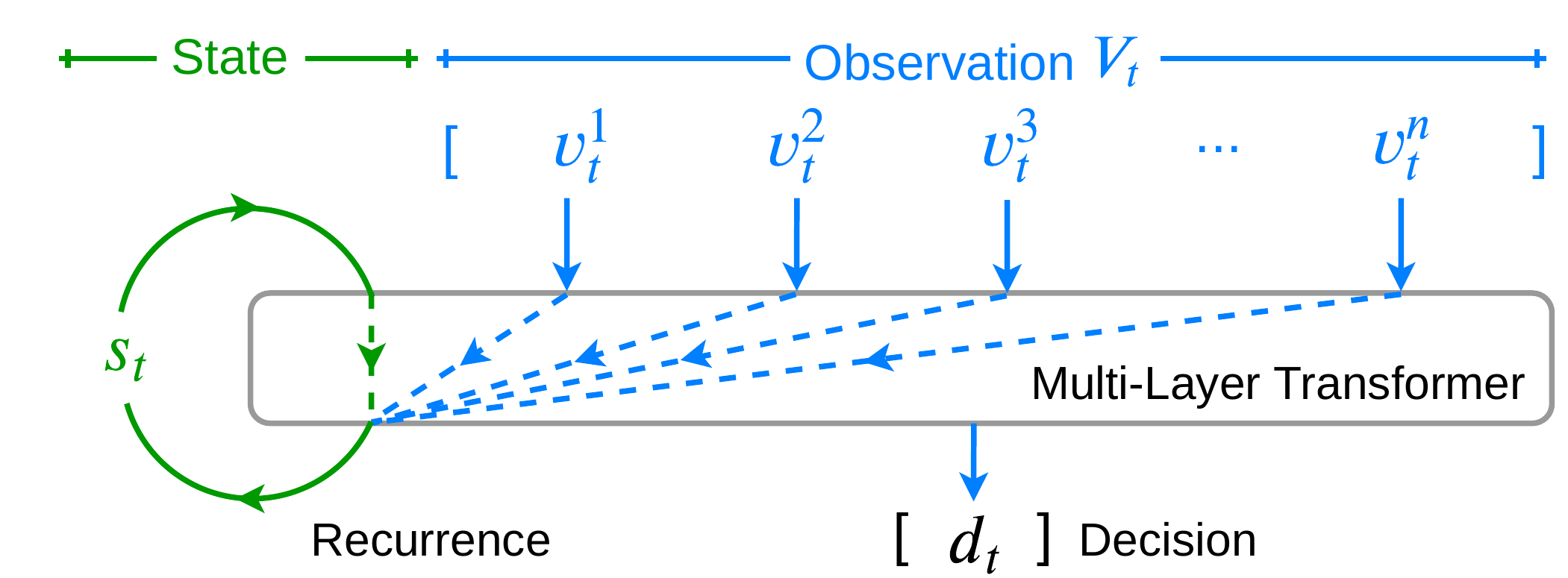}
  \caption{Recurrent multi-layer Transformer for addressing partially observable inputs. A state token is defined along with the input sequence. At each time step, a new state representation $\boldsymbol{s}_{t}$ will be generated based on the new observation. Meanwhile, the past information will help inferring a new decision $d_{t}$.}
  \label{fig:title}
  \vspace{-1em}
\end{figure}

However, an essential difference between VLN and other vision-and-language tasks is that VLN can be considered as a partially observable Markov decision process, in which future observations are dependent on the agent's current state and action. Meanwhile, at each navigational step, the visual observation only corresponds to partial instruction, requiring the agent to keep track of the navigation progress and correctly localise the relevant sub-instruction to gain useful information for decision making. Another difficulty of applying V\&L BERT for VLN is the high demand on computational power; since the navigational episode could be very long, performing self-attention on a long visual and textual sequence at each time step will cost an excessive amount of (GPU) memory during training.

To address the aforementioned problems, we propose a recurrent vision-and-language BERT for navigation, or simply \vlnbert. Instead of employing large-scale datasets for pre-training which usually require thousands of GPU hours, the aim of this work is to allow the learning of VLN to adequately benefit from pre-trained V\&L BERT. Based on the previously proposed V\&L BERT models, we implement a recurrent function in their original architecture (Fig.~\ref{fig:title}) to model and leverage the history-dependent state representations, without explicitly defining a memory buffer~\cite{zhu2020babywalk} or applying any external recurrent modules such as an LSTM~\cite{hochreiter1997long}. To reduce the memory consumption, we control the self-attention to consider the language tokens as keys and values but not queries during navigation, which is similar to the cross-modality encoder in LXMERT~\cite{tan2019lxmert}. Such design greatly reduces the memory usage so that the entire model can be trained on a single GPU without performance degeneration. Furthermore, as in the original V\&L BERT, our proposed model has the potential of multi-task learning, it is able to address other vision and language problems along with the navigation task. 

We employ two datasets to evaluate the performance of our \vlnbert, R2R~\cite{anderson2018vision} and REVERIE~\cite{qi2020reverie}. The chosen datasets are different in terms of the provided visual clues, the instructions and the goal. Our agent, initialised from a pre-trained V\&L BERT and fine-tuned on the two datasets, achieves state-of-the-art results. We also initialise our model with the PREVALENT \cite{hao2020towards}, a LXMERT-like model pre-trained for VLN. On the test split of R2R \cite{anderson2018vision}, it improves the Success Rate absolutely by 8\% and achieves 57\% Success weighted by Path Length (SPL). For the remote referring expression task in REVERIE \cite{qi2020reverie}, our agent obtains 23.99\% navigation SPL and 13.51\% Remote Grounding SPL. These results indicate the strong generalisation ability of our proposed \vlnbert~ as well as the potential of using it for merging the learning of VLN with other vision and language tasks.

\begin{figure*}[t]
  \centering
  \includegraphics[width=\textwidth]{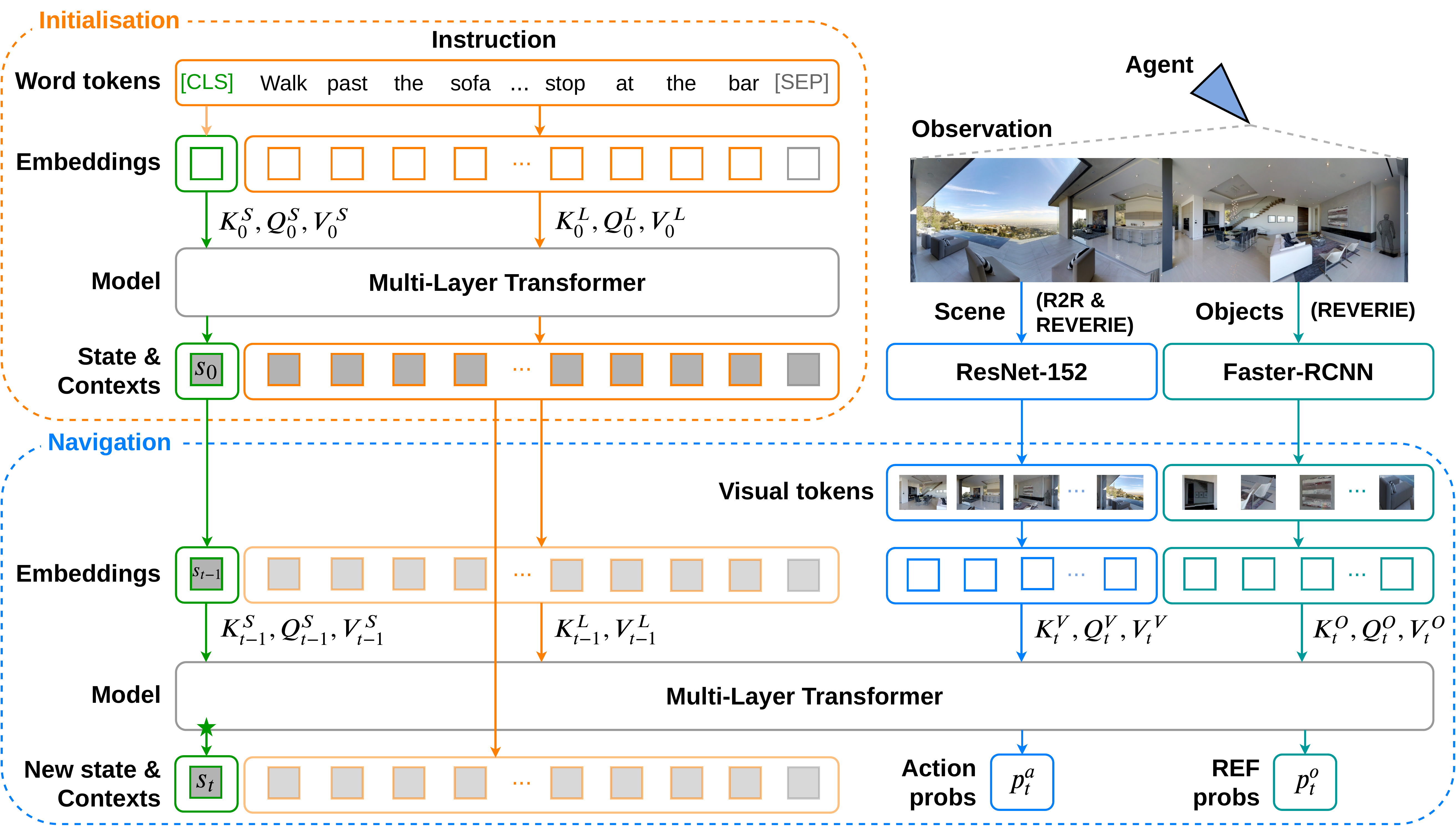}
  \caption{Schematics of the Recurrent Vision-and-Language BERT. At the initialisation stage, the entire instruction is encoded by a multi-layer Transformer, where the output feature of the \texttt{[CLS]} token serves as the initial state representation of the agent. During navigation, the concatenated sequence of state, encoded language and new visual observation is fed to the same Transformer to obtain the updated state and decision probabilities. The updated state and the language encoding from initialisation will be fused and applied as input at the next time step. The green star (\star) indicates the \textit{cross-modal matching} (Eq.~\ref{eqn:matching}) and the \textit{past decision encoding} (Eq.~\ref{eqn:stateaction}) in \textit{State Refinement}.} % , which are omitted here for simplicity.
  \label{fig:vln_bert}
  \vspace{-1em}
\end{figure*}

\section{Related Work} 
\label{sec:relatedwork}

\paragraph{Vision-and-Language Navigation}
Learning navigation with visual-linguistic clues has drawn significant research interests. The recent R2R \cite{anderson2018vision} and Touchdown \cite{chen2019touchdown} datasets introduce human natural language as guidance and apply photo-realistic environments for navigation. Following these work, dialog-based navigation such as CVDN \cite{thomason2020vision}, VNLA \cite{nguyen2019vision} and HANNA \cite{nguyen2019help}, navigation for localising a remote object such as REVERIE \cite{qi2020reverie}, VLN in continuous environment \cite{krantz2020navgraph}, and multilingual navigation with spatial-temporal grounding such as RxR \cite{anderson2020rxr} have been proposed for further research. 

One crucial challenge in VLN is to understand the visual-textual correspondence for decision making. To achieve this, Self-Monitoring~\cite{ma2019self} and RCM \cite{wang2019reinforced} adopt cross-modal attention to highlight the relevant observations and instruction at each step. Speaker-Follower \cite{fried2018speaker} and EnvDrop \cite{tan2019learning} learn on the augmented training data via a self-supervised manner. FAST \cite{ke2019tactical} resorts self-correction navigation, while
APS \cite{fu2020counter} samples adversarial paths for training to enhance the model's ability to generalise. AuxRN \cite{zhu2020vision} applies several auxiliary losses to learn comprehensive representations, Qi \etal \cite{qi2020object} and Wang \etal \cite{wang2020soft} also design loss functions to encourage the agent to follow the instructions to take the shortest paths. More recently, Hong \etal \cite{hong2020graph} propose a graph network to model the intra- and inter-modal relationships among the contextual and visual clues. The great improvements achieved by these methods encourage researchers to explore simpler and more powerful visiolinguistic learning network for VLN.

\vspace{-15pt}
\paragraph{Visual BERT Pre-Training} 
Following the success of pre-trained BERT on a wide range of natural language processing tasks \cite{devlin2019bert}, the model has been extended to process visual tokens and to pre-train on large-scale image/video-text pairs for learning generic visual-linguistic representations. Previous research introduce two-stream BERT models which encode texts and images separately, and fuse the two modalities in a later stage \cite{lu2019vilbert,tan2019lxmert}, as well as one-stream BERT models which directly perform inter-modal grounding \cite{chen2020uniter,li2020unicoder,li2019visualbert,li2020oscar,su2019vl}. Although video BERT approaches have been proposed to learn the correspondence between texts and video frames \cite{li2020hero,luo2020univilm,sun2019videobert,yang2020bert}, we are the first to integrate recurrence into BERT to learn partially-observable and temporal-dependent inputs.
In terms of VLN pre-training, PRESS fine tunes a pre-trained language BERT to encode instructions \cite{li2019robust}, PREVALENT trains a V\&L BERT on a large amount of image-text-action triplets from scratch to learn navigation-oriented textual representations \cite{hao2020towards}, and VLN-BERT \cite{majumdar2020improving} fine-tunes a ViLBERT~\cite{lu2019vilbert} on instruction-trajectory pairs to measure their compatibility in beam search setting. Unlike all previous work, our \vlnbert~ can augment various V\&L BERT models with recurrent function, it is a navigator network by itself that can be directly trained for navigation.

\vspace{-15pt}
\paragraph{V\&L Multi-Task Learning}
Instead of building a monolithic model for different V\&L tasks, numbers of previous work explore multi-task learning with a unified model for utilising the common and the complementary knowledge to reduce the domain gap \cite{li2018visual,nguyen2019multi,pramanik2019omninet,shuster2019dialogue,lu2019vilbert}.
Very recently, 12-in-1 \cite{lu202012} trains a single ViLBERT \cite{lu2019vilbert} on 12 different datasets across four categories of V\&L tasks, including visual question answering, referring expressions, multi-modal verification and caption-based image retrieval.
In Vision-and-Language Navigation, Wang \etal~\cite{wang2020environment} propose a multitask navigation model to address the R2R navigation \cite{anderson2018vision} and the Navigation from Dialog History (NDH) \cite{thomason2020vision} problems. Comparing to previous methods, our \vlnbert~uses a single network to address the navigation task and the remote referring expression task seamlessly in REVERIE \cite{qi2020reverie}.

\section{Proposed Model} 
\label{sec:rvlnbert}

In this section, we first define the vision-and-language navigation task, then we revisit the BERT model \cite{devlin2019bert} and present the architecture of our proposed \vlnbert~.

\subsection{VLN Background}
The problem of VLN can be formulated as follows: Given a natural language instruction $\boldsymbol{U}$ which contains a sequence of words, at each time step $t$, the agent observes the environment and infers an action $a_{t}$ that transfers the agent from state $\boldsymbol{s}_{t}$ to a new state $\boldsymbol{s}_{t+1}$. The state consists of the navigational history and the current spatial position defined by a triplet $\langle \boldsymbol{C}_{t},\theta_{t},\phi_{t} \rangle$, where $\boldsymbol{C}_{t}$ is a viewpoint on the pre-defined connectivity graph of the environment \cite{anderson2018vision}, and $\theta_{t}$ and $\phi_{t}$ are the angles of heading and elevation, respectively. The agent needs to execute a sequence of actions to navigate on the connectivity graph and eventually decides to \texttt{stop} at the target position to complete the task.

\subsection{Revisit BERT}
Bidirectional Encoder Representations from Transformers (BERT) \cite{devlin2019bert} is a multi-layer Transformer architecture~\cite{vaswani2017attention} designed to pre-train deep bidirectional language representations.
Each layer of the Transformer encodes the language features from the previous layer $\boldsymbol{X}_{l-1}\in\mathbb{R}^{I\times{hd_{h}}}$ with multi-head self-attention to capture the dependencies among the $I$ words in the sentence, and applies a residual feed-forward network to process the output features.

Formally, the $k$-th attention head at the $l$-th layer performs self-attention over $\boldsymbol{X}_{l-1}$ as
\begin{gather}
\boldsymbol{Q} = \boldsymbol{X}_{l-1}\boldsymbol{W}^{Q}_{l,k}, \boldsymbol{K} = \boldsymbol{X}_{l-1}\boldsymbol{W}^{K}_{l,k}, \boldsymbol{V} = \boldsymbol{X}_{l-1}\boldsymbol{W}^{V}_{l,k} \\
\boldsymbol{H}_{l,k} = 
% \text{SelfAttn}(\boldsymbol{Q}, \boldsymbol{K}, \boldsymbol{V}) =
\text{Softmax}\left(\frac{\boldsymbol{Q}\boldsymbol{K}^\top}{\sqrt{d_h}}\right)\boldsymbol{V}
\label{eqn:selfattn}
\end{gather}
where $\boldsymbol{W}^{Q}$, $\boldsymbol{W}^{K}$ and $\boldsymbol{W}^{V}\in\mathbb{R}^{hd_{h}\times{d}_{h}}$ are learnable linear projections\footnote{All $\boldsymbol{W}$ in this section denotes learnable linear projections} specifically for queries, keys and values; $d_{h}$ is the hidden dimension of the network.
The outputs from all the attention heads will be concatenated and projected onto the same dimension as the input as
\begin{equation}
\boldsymbol{H}_{l} = \left[\boldsymbol{H}_{l,1};...;\boldsymbol{H}_{l,h}\right]\boldsymbol{W}^{O}_{l}
\label{eqn:multihead}
\end{equation}
where $h$ is the total number of heads, $[;]$ denotes concatenation and $\boldsymbol{W}^{O}\in\mathbb{R}^{hd_{h}\times{hd_{h}}}$ is a learned linear projection.
Finally, the output of layer $l$ is formulated by
\begin{align}
\boldsymbol{H}'_{l} &= \text{LayerNorm}(\boldsymbol{H}_{l}+\boldsymbol{X}_{l-1}) \\
\boldsymbol{X}'_{l} &= \text{ReLU}(\boldsymbol{H}'_{l}\boldsymbol{W}^{F_1}_{l})\boldsymbol{W}^{F_2}_{l} \\
\boldsymbol{X}_{l} &= \text{LayerNorm}(\boldsymbol{H}'_{l}+\boldsymbol{X}'_{l})
\label{eqn:layerout}
\end{align}
% where GELU is the Gaussian Error Linear Unit activation function \cite{hendrycks2016gaussian}
where ReLU is the Rectified Linear Unit activation function and LayerNorm is layer normalisation \cite{ba2016layer}.

Based on this architecture, BERT has been extended to V\&L BERT \cite{chen2020uniter,li2020unicoder, li2019visualbert,li2020oscar,lu2019vilbert,su2019vl,tan2019lxmert}, which takes the concatenation of language tokens and visual tokens as input, and pre-trains on image-text corpus to learn generic visiolinguistic representations.

% % ---------------------------------------

\subsection{Recurrent VLN BERT}
The idea of our \vlnbert~ can be adapted to a wide range of Transformer-based networks. In this section we apply the recently proposed one-stream V\&L BERT model OSCAR \cite{li2020oscar} for demonstration. We modify the model to enable the learning of navigation and the associated referring expression (REF) task. As shown in Fig.~\ref{fig:vln_bert}, at each time step, the network takes four sets of tokens as input; the previous state token $\boldsymbol{s}_{t-1}$, the language tokens $\boldsymbol{X}$, the visual tokens $\boldsymbol{V}_{\!t}$ for scene, and the visual tokens $\boldsymbol{O}_{\!t}$ for objects (only in REVERIE~\cite{qi2020reverie}). Then, it performs self-attention over these cross-modal tokens to capture the textual-visual correspondence for inferring the action probabilities $\boldsymbol{p}^{a}_{t}$ and the object grounding probabilities $\boldsymbol{p}^{o}_{t}$ (only in REVERIE~\cite{qi2020reverie}):
\begin{equation}
\boldsymbol{s}_{t},\boldsymbol{p}^{a}_{t}, \boldsymbol{p}^{o}_{t} = \text{VLN}\CircleArrowright\text{BERT}(\boldsymbol{s}_{t-1},\boldsymbol{X},\boldsymbol{V}_{\!t},\boldsymbol{O}_{\!t})
\label{eqn:vlnrbert}
\end{equation}

\vspace{-14pt}
\paragraph{Language Processing}
At initialisation ($t{=}0$), a sequence of words consisting of the classification token \texttt{[CLS]}, the language tokens of the instruction $\boldsymbol{U}$ and the separation token \texttt{[SEP]} will be fed into \vlnbert, where \texttt{[CLS]} and \texttt{[SEP]} are pre-defined in BERT models.
In the pre-training of OSCAR~\cite{li2020oscar}, the \texttt{[CLS]} token is applied for aggregating relevant visiolinguistic clues from the input sequence for contrastive learning. 
% In previous V\&L BERT, the \texttt{[CLS]} token is trained for gathering relevant visiolinguistic clues from the input sequence, and it is usually applied for downstream classification tasks. 
% We argue that the function of \texttt{[CLS]} can be adopted to represent the agent's state in VLN. 
Here, we defined the embedded \texttt{[CLS]} token as the initial state representation $\boldsymbol{s}_{0}$, to inherit such function — initialise an agent’s state which is aware of the entire navigation task.
\begin{equation}
\boldsymbol{s}_{0},\boldsymbol{X}=\text{VLN}\CircleArrowright\text{BERT}(\texttt{[CLS]},\boldsymbol{U},\texttt{[SEP]})
\label{eqn:init}
\end{equation}
During navigation steps ($t{>}0$), unlike the state token $\boldsymbol{s}_{t}$ or the visual tokens $\boldsymbol{V}_{\!t}$ and $\boldsymbol{O}_{\!t}$ which performs self-attention with respect to the entire input sequence, the language tokens $\boldsymbol{X}$ only serve as the keys and values in the Transformer. We consider the language tokens produced by the model at the initialisation step as a deep representation of the instruction which does not need to be further encoded in later steps. Not updating the language features also save a huge amount of computational resources since the instruction and the trajectory can be long in VLN problems.

\vspace{-14pt}
\paragraph{Vision Processing}
At each navigation step ($t{>}0$), the agent makes new visual observation in the environment and uses the visual clues to assist navigation. To process the visual clues, the network first projects the image features of views at the navigable directions $\boldsymbol{I}^{v}_{t}$ to the same space as the BERT token as $\boldsymbol{V}_{\!t}{=}\boldsymbol{I}^{v}_{t}\boldsymbol{W}^{I^{v}}$. Then, the visual tokens will be concatenated with the state token and the language tokens, and fed into the model.

In terms of the remote REF task \cite{qi2020reverie}, we simply consider the object features $\boldsymbol{I}^{o}_{t}$ as additional visual tokens in the input sequence. Similarly, the features will be projected onto the token space as $\boldsymbol{O}_{t}{=}\boldsymbol{I}^{o}_{t}\boldsymbol{W}^{I^{o}}$ and fed into the model. The object clues can provide valuable information about the important landmarks on the path, which could be very helpful to the navigation with high-level instructions \cite{qi2020reverie}.

% -----------------------------------

\vspace{-14pt}
\paragraph{State Representation} 
We formulate the agent's state at each time step $\boldsymbol{s}_{t}$ as the summary of all textual and visual clues that the agent collects, as well as all decisions that the agent makes until the current viewpoint. Instead of explicitly defining a memory buffer \cite{zhu2020babywalk} or implementing an additional recurrent network \cite{hochreiter1997long} to store the past experiences, our model relies on BERT's original architecture to recognise time-dependent inputs, and recurrently updates $\boldsymbol{s}_{0}$ from initialisation to represent the state.
At each navigation step, the state representation is used as the leading input token of the entire textual-visual sequence. It then performs inter-modal self-attention in \vlnbert~with other tokens to update its content and becomes the leading token of the input at the next step, in an autoregressive way.

\vspace{-14pt}
\paragraph{State Refinement} 
Unlike most of the V\&L BERT models which apply the output feature of the \texttt{[CLS]} token for classification, our state is not directly used for inferring a decision (see following \textit{Decision Making} subsection), which means, the vanilla state representation is not explicitly enforced to capture the most important language and visual features. To address this issue, our model matches the raw textual and visual tokens, and feeds the output to the state representation. Formally, let $\boldsymbol{Q}^{s}_{l,k}$ and $\boldsymbol{K}^{x}_{l,k}$ be the state and textual tokens at head $k$ of the final ($l{=}12$) layer of \vlnbert, the attention scores over the textual tokens can be expressed as:
\begin{equation}
\boldsymbol{A}^{s,x}_{l,k} = \frac{\boldsymbol{Q}^{s}_{l,k}{\boldsymbol{K}^{x}_{l,k}}^{\!\top}}{\sqrt{d_h}}
\label{eqn:lang_attn}
\end{equation}
Then, we average the scores over all the attention heads ($K{=}12$) and apply a Softmax function to get the overall state-language attention weights as:
\begin{equation}
\widetilde{\boldsymbol{A}}^{s,x}_{l} = \text{Softmax}(\widebar{\boldsymbol{A}}^{s,x}_{l})= \text{Softmax}\left(\frac{1}{K}\sum^{K}_{k=1}\boldsymbol{A}^{s,x}_{l,k}\right)
\label{eqn:lang_weights}
\end{equation}
Similarly, the visual attention scores $\boldsymbol{A}^{s,v}_{l,k}$ and weights $\widetilde{\boldsymbol{A}}^{s,v}_{l}$ can be obtained.
Now, we perform a weighted sum over the input textual tokens and visual tokens respectively to obtain the weighted raw features as:
\begin{equation}
\boldsymbol{F}^{x}_{t} = \widetilde{\boldsymbol{A}}^{s,x}_{l}\boldsymbol{X}
\quad\text{and}\quad
\boldsymbol{F}^{v}_{t} = \widetilde{\boldsymbol{A}}^{s,v}_{l}\boldsymbol{V}_{\!t}
\label{eqn:weight_raw}
\end{equation}
We then enforce a \textbf{cross-modal matching} between the raw textual and visual features via element-wise product and send such information to the agent's state as:
\begin{equation}
\boldsymbol{s}^{f}_{t}=\left[\boldsymbol{s}^{r}_{t};\boldsymbol{F}^{x}_{t} \odot \boldsymbol{F}^{v}_{t}\right]\boldsymbol{W}^{r}
\label{eqn:matching}
\end{equation}
where $\boldsymbol{s}^{r}_{t}$ is the output state features at the final layer. Notice that in REVERIE \cite{qi2020reverie}, only the visual features is sent to the state representation, i.e., Eq.~\ref{eqn:matching} becomes $\boldsymbol{s}^{f}_{t}=\left[\boldsymbol{s}^{r}_{t};\boldsymbol{F}^{v}_{t}\right]\boldsymbol{W}^{r}$. This is because the navigational instructions in REVERIE are high-level, hence performing step-wise matching between the raw textual and visual features is less valuable.

Finally, past decisions are important for the agent to keep track of the navigation progress, our network records the new decision by feeding the directional features of the selected action $\boldsymbol{a}_{t}$ into the state token as:
\begin{equation}
\boldsymbol{s}_{t}=\left[\boldsymbol{s}^{f}_{t};\boldsymbol{a}_{t}\right]\boldsymbol{W}^{s}
\label{eqn:stateaction}
\end{equation}
where $\boldsymbol{s}_{t}$ is the new representation of the agent's state at time step $t$.

% ---------------------------------------

\paragraph{Decision Making \label{decision}}
Many previous VLN agents apply an inner product between the state representation and the visual features at candidate directions to evaluate the state-vision correspondence, and choose a direction with the highest matching score to navigate \cite{ma2019self, tan2019learning}. We find that the BERT network can nicely perform such scoring because it is fully built upon the inner product based soft-attention. Inspired by the method of predicting alignment between regions and phrases in VisualBERT~\cite{li2019visualbert}, we directly apply the mean attention weights of the visual tokens over all the attention heads in the last layer, with respect to the state, as the action probabilities, simply $\boldsymbol{p}^{a}_{t}=\widetilde{\boldsymbol{A}}^{s,v}_{l}$ (as defined in Eq.~\ref{eqn:lang_weights}).
As for the remote referring expression task \cite{qi2020reverie}, our agent uses the same method to select an object. The selection probabilities can be expressed as $\boldsymbol{p}^{o}_{t}=\widetilde{\boldsymbol{A}}^{s,o}_{l}$, where $\widetilde{\boldsymbol{A}}^{s,o}_{l}$ is the mean attention weights for all candidate objects. We refer the \textit{Appendix} \S B.2 for more details.

% --------------------------------------------------------------------

\subsection{Training}
We train our network with a mixture of reinforcement learning (RL) and imitation learning (IL) objectives. We apply A2C \cite{mnih2016asynchronous} for RL, in which the agent samples an action according to $\boldsymbol{p}^{a}_{t}$ and measures the advantage $\boldsymbol{A}_{t}$ at each step (we refer the \textit{Appendix} \S B.5 for more details about RL.). In IL, our agent navigates on the ground-truth trajectory by following teacher actions and calculates a cross-entropy loss for each decision. Formally, we minimise the navigation loss function, expressed for each given sample, as
\begin{equation}
\mathcal{L} = -\sum_{t} a^{s}_{t}\text{log}\left(p^{a}_{t}\right)A_{t} - \lambda \sum_{t} a^{*}_{t}\text{log}\left(p^{a}_{t}\right)
\label{eqn:objective}
\end{equation}
where $a^{s}_{t}$ is the sampled action and $a^{*}_{t}$ is the teacher action. Here $\lambda$ is a coefficient for weighting the IL loss. In REVERIE \cite{qi2020reverie}, we applied an additional cross-entropy term $\sum_{t} o^{*}_{t}\text{log}(p^{o}_{t})$ to learn object grounding.

\subsection{Adaptation}
We initialise the parameters of \vlnbert~ from OSCAR \cite{li2020oscar} pre-trained without object tags. Although OSCAR is trained on regional features, we find that it is also compatible with the grid features of the entire scene. 
When adapting to the LXMERT-like \cite{tan2019lxmert} model in PREVALENT~\cite{hao2020towards}, we remove the language branch in the cross-modality encoder and concatenate the state token with the visual tokens for self-attention (see \textit{Appendix} \S B.3 for schematics). We also remove the entire downstream network EnvDrop \cite{tan2019learning}, including the Speaker and the environmental dropout, and then directly fine-tune the model pre-trained by PREVALENT for navigation.

\section{Experiments} 
\label{sec:experiments}

% \subsection{Setup}

% % ----------------------------------------------

\begin{table*}[t]
  \begin{center}
  \resizebox{0.90\textwidth}{!}{
  \begin{tabular}{l|rrrr|rrrr|rrrr}
    \hline \hline
    \multicolumn{1}{c}{\multirow{2}{*}{Methods}} & \multicolumn{4}{|c}{R2R Validation Seen} & \multicolumn{4}{|c}{R2R Validation Unseen} & \multicolumn{4}{|c}{R2R Test Unseen} \\
    \cline{2-13} &
    \multicolumn{1}{c}{TL} & \multicolumn{1}{c}{NE$\downarrow$} & \multicolumn{1}{c}{SR$\uparrow$} & \multicolumn{1}{c|}{SPL$\uparrow$} & \multicolumn{1}{c}{TL} & \multicolumn{1}{c}{NE$\downarrow$} & \multicolumn{1}{c}{SR$\uparrow$} & \multicolumn{1}{c|}{SPL$\uparrow$} & \multicolumn{1}{c}{TL} & \multicolumn{1}{c}{NE$\downarrow$} & \multicolumn{1}{c}{SR$\uparrow$} & \multicolumn{1}{c}{SPL$\uparrow$} \Tstrut\\
    \hline \hline
    Random     & 9.58  & 9.45 & 16   & -    & 9.77  & 9.23 & 16   & -    & 9.89  & 9.79 & 13 & 12 \\
    Human      & -     & -    & -    & -    & -     & -    & -    & -    & 11.85 & 1.61 & 86 & 76 \\
    \hline
    Seq2Seq-SF \cite{anderson2018vision}
        & 11.33 & 6.01 & 39 & -    & 8.39  & 7.81 & 22 & -    & 8.13  & 7.85 & 20 & 18 \\
    Speaker-Follower \cite{fried2018speaker}
        & -     & 3.36 & 66 & -    & -     & 6.62 & 35 & -    & 14.82 & 6.62 & 35 & 28 \\
    SMNA \cite{ma2019self}
        & -     & 3.22 & 67 & 58 & -     & 5.52 & 45 & 32 & 18.04 & 5.67 & 48 & 35 \\
    RCM+SIL (train) \cite{wang2019reinforced}
        & 10.65 & 3.53 & 67 & -    & 11.46 & 6.09 & 43 & -    & 11.97 & 6.12 & 43 & 38 \\
    PRESS \cite{li2019robust} $\dag$
        & 10.57 & 4.39 & 58 & 55 & 10.36 & 5.28 & 49 & 45 & 10.77 & 5.49 & 49 & 45 \\
    FAST-Short \cite{ke2019tactical}
        & -     & -    & -    & -    & 21.17 & 4.97 & 56 & 43 & 22.08 & 5.14 & 54 & 41 \\
    EnvDrop \cite{tan2019learning}
        & 11.00 & 3.99 & 62 & 59 & 10.70 & 5.22 & 52 & 48 & 11.66 & 5.23 & 51 & 47 \\
    AuxRN \cite{zhu2020vision}
        & -     & 3.33 & 70 & \textbf{67} & -     & 5.28 & 55 & 50 & -     & 5.15 & 55 & 51 \\
    PREVALENT \cite{hao2020towards} $\dag$
        & 10.32 & 3.67 & 69 & 65 & 10.19 & 4.71 & 58 & \textbf{53} & 10.51 & 5.30 & 54 & 51 \\
    RelGraph \cite{hong2020graph} 
        & 10.13 & 3.47 & 67 & 65 & 9.99 & 4.73 & 57 & \textbf{53} & 10.29 & 4.75 & 55 & 52 \\
    \hline
    Ours (no init. OSCAR) 
        & 9.78 & 3.92 & 62 & 59 & 10.31 & 5.10 & 50 & 46 & 11.15 & 5.45 & 51 & 47\\
    Ours (init. OSCAR)
        & 10.79 & \textbf{3.11} & \textbf{71} & \textbf{67} & 11.86 & \textbf{4.29} & \textbf{59} & \textbf{53} & 12.34 & \textbf{4.59} & \textbf{57} & \textbf{53} \\
    % \hline
    Ours (init. PREVALENT)
        & 11.13 & \high{2.90} & \high{72} & \high{68} & 12.01 & \high{3.93} & \high{63} & \high{57} & 12.35 & \high{4.09} & \high{63} & \high{57} \\
    \hline \hline
  \end{tabular}}
\end{center}
\caption{Comparison of agent performance on R2R in single-run setting. $\dag$ work that applies pre-trained BERT for language encoding.}
\label{tab:table_r2r_result}
\end{table*}

% % ----------------------------------------------

\begin{table*}[t]
\begin{center}
\resizebox{\linewidth}{!}{
\begin{tabular}{l|rrrrrr|rrrrrr|rrrrrr}
\hline \hline
\multicolumn{1}{c}{\multirow{3}{*}{Methods}}  & \multicolumn{6}{|c|}{REVERIE Validation Seen} &\multicolumn{6}{c|}{REVERIE Validation Unseen} & \multicolumn{6}{c}{REVERIE Test Unseen} \Tstrut\\
\cline{2-19}& \multicolumn{4}{c|}{Navigation}  & \multicolumn{1}{c}{\multirow{2}{*}{RGS$\uparrow$}}&\multicolumn{1}{c|}{\multirow{2}{*}{RGSPL$\uparrow$}} & \multicolumn{4}{c|}{Navigation}  & \multicolumn{1}{c}{\multirow{2}{*}{RGS$\uparrow$}}&\multicolumn{1}{c|}{\multirow{2}{*}{RGSPL$\uparrow$}} & \multicolumn{4}{c|}{Navigation}   & \multicolumn{1}{c}{\multirow{2}{*}{RGS$\uparrow$}}&\multicolumn{1}{c}{\multirow{2}{*}{RGSPL$\uparrow$}} \\
\cline{2-5} \cline{8-11} \cline{14-17}  & \multicolumn{1}{c}{SR$\uparrow$} & \multicolumn{1}{c}{OSR$\uparrow$} & \multicolumn{1}{c}{SPL$\uparrow$}  & \multicolumn{1}{c|}{TL} &  &  & \multicolumn{1}{c}{SR$\uparrow$} & \multicolumn{1}{c}{OSR$\uparrow$} & \multicolumn{1}{c}{SPL$\uparrow$} &\multicolumn{1}{c|}{TL} & & & \multicolumn{1}{c}{SR$\uparrow$} & \multicolumn{1}{c}{OSR$\uparrow$} & \multicolumn{1}{c}{SPL$\uparrow$} & \multicolumn{1}{c|}{TL} &  & \Tstrut\\
\hline \hline
Random &2.74 &8.92& 1.91 & \multicolumn{1}{c|}{11.99} & 1.97 & 1.31 &1.76& 11.93 &1.01& \multicolumn{1}{c|}{10.76}  & 0.96 & 0.56 &  2.30  &8.88 & 1.44&  \multicolumn{1}{c|}{10.34} &1.18 & 0.78 \\
Human & -- & --  & -- & \multicolumn{1}{r|}{--} & -- & -- & -- & --  & --  & \multicolumn{1}{r|}{--} & -- & -- & 81.51 & 86.83  & 53.66 & \multicolumn{1}{r|}{21.18} & 77.84 &  51.44 \\
\hline
Seq2Seq-SF \cite{anderson2018vision} &29.59&35.70 & 24.01 & \multicolumn{1}{r|}{12.88}  &18.97& 14.96 & 4.20& 8.07 & 2.84   & \multicolumn{1}{r|}{11.07}  & 2.16 & 1.63 & 3.99 & 6.88  & 3.09 & \multicolumn{1}{r|}{10.89} & 2.00 & 1.58 \\
RCM \cite{wang2019reinforced} & 23.33 & 29.44 & 21.82& \multicolumn{1}{r|}{10.70} & 16.23 & 15.36 & 9.29 & 14.23 & 6.97 &\multicolumn{1}{r|}{11.98} & 4.89 & 3.89 & 7.84 & 11.68 & 6.67 & \multicolumn{1}{r|}{10.60} & 3.67 & 3.14\\
SMNA \cite{ma2019self} & 41.25& 43.29  & 39.61& \multicolumn{1}{r|}{7.54}   & 30.07 &  28.98 & 8.15 & 11.28 &6.44 & \multicolumn{1}{r|}{9.07} & 4.54& 3.61 & 5.80& 8.39 & 4.53 &\multicolumn{1}{r|}{9.23}   & 3.10& 2.39 \\
FAST-Short \cite{ke2019tactical} & 45.12& 49.68 &40.18& \multicolumn{1}{r|}{13.22}  &31.41 & 28.11 & 10.08 & 20.48 & 6.17 & \multicolumn{1}{r|}{29.70}  & 6.24 & 3.97 & 14.18 & 23.36 & 8.74 & \multicolumn{1}{r|}{30.69}  & 7.07 & 4.52 \\
FAST-MATTN \cite{qi2020reverie} & \textbf{50.53} & \high{55.17} & \textbf{45.50} & \multicolumn{1}{r|}{{16.35}} & \textbf{31.97} & \textbf{29.66} & 14.40 & 28.20 & 7.19 & \multicolumn{1}{r|}{{45.28}}  & 7.84 & 4.67 & 19.88 & \textbf{30.63} & 11.61 & \multicolumn{1}{r|}{{39.05}} & 11.28 & 6.08 \\
\hline
Ours (no init. OSCAR) & 39.56 & 42.09 & 36.29 & \multicolumn{1}{r|}{{12.06}} & 21.15 & 19.39 & \textbf{25.76} & \textbf{29.28} & \textbf{22.16} & \multicolumn{1}{r|}{{14.52}} & 11.62 & 9.87 & 18.52 & 20.18 & 15.47 & \multicolumn{1}{r|}{{14.09}} & 8.80 & 7.29 \\
Ours (init. OSCAR) & 39.85 & 41.32 & 35.86 & \multicolumn{1}{r|}{{12.85}} & 24.46 & 22.28 & 25.53 & 27.66 & 21.06 & \multicolumn{1}{r|}{{14.35}} & \textbf{14.20} & \textbf{12.00} & \textbf{24.62} & 26.67 & \textbf{19.48} & \multicolumn{1}{r|}{{14.88}} & \textbf{12.65} & \textbf{10.00} \\
\hline
Ours (init. PREVALENT) & \high{51.79} & \textbf{53.90} & \high{47.96} & \multicolumn{1}{r|}{{13.44}} & \high{38.23} & \high{35.61} & \high{30.67} & \high{35.02} & \high{24.90} & \multicolumn{1}{r|}{{16.78}} & \high{18.77} & \high{15.27} & \high{29.61} & \high{32.91} & \high{23.99} & \multicolumn{1}{r|}{{15.86}} & \high{16.50} & \high{13.51} \\
\hline \hline
\end{tabular}}
\end{center}
\caption{Comparison of agent performance of navigation and remote referring expression on REVERIE.}
\label{tab:reverie}
\vspace{-10pt}
\end{table*}

\paragraph{Implementation Details}
All experiments are conducted on a single NVIDIA 2080Ti GPU, the learning rate is fixed to $10^{-5}$ throughout the training and AdamW optimiser \cite{loshchilov2018decoupled} is applied. For R2R, we train the agent directly on the mixture of the original training data and the augmented data from PREVALENT \cite{hao2020towards}, the batch size\footnote{Half for RL and half for IL in each iteration, corresponding to the first and the second term in Eq.~\ref{eqn:objective}, respectively. \label{footnote_bs}} is set to 16 and the network is trained for 300,000 iterations. For REVERIE, we use batch size$^{\ref{footnote_bs}}$ 8 and train the agent for 200,000 iterations. Images in the environments are encoded by a ResNet-152 \cite{he2016deep} pre-trained on Places365 \cite{zhou2017places}, and objects are encoded by a Faster-RCNN \cite{ren2015faster} pre-trained on the Visual Genome \cite{krishna2017visual}. Early stopping is applied when the training saturates, the model which achieves the highest SPL in validation unseen split is adopted for testing.

\vspace{-5pt}
\paragraph{Evaluation Metrics} 
We apply the standard metrics employed by previous works to evaluate the performance.

R2R \cite{anderson2018vision} considers Trajectory Length (TL): the average path length in meters, Navigation Error (NE): the average distance between agent's final position and the target in meters, Success Rate (SR): the ratio of stopping within 3 meters to the target, and Success weighted by the normalised inverse of the Path Length (SPL) \cite{anderson2018evaluation}.

REVERIE \cite{qi2020reverie} defines Success Rate (SR) as the ratio of stopping at a viewpoint where the target object is visible (in panorama), and considers the corresponding SPL. It also employ Oracle Success Rate (OSR): the ratio of having a viewpoint along the trajectory where the target object is visible, Remote Grounding Success Rate (RGS): the ratio of grounding to the correct objects when stopped, and RGSPL, which weights RGS by the trajectory length.

% % ----------------------------------------------

\subsection{Main Results}

\paragraph{Comparison with SoTA}
Results in Table~\ref{tab:table_r2r_result} compare the single-run (greedy search, no pre-exploration \cite{wang2019reinforced}) performance of different agents on the R2R benchmark. Our proposed \vlnbert~ initialised from OSCAR \cite{li2020oscar} (init. OSCAR) performs better than previous methods across all the dataset splits. Comparing to a randomly initialised network (no init. OSCAR), the large performance degeneration suggests that the pre-trained general vision-linguistic knowledge significantly benefits the learning of navigation. The model initialised from PREVALENT \cite{hao2020towards}, pre-trained especially for VLN, further improves the agent's performance, achieving 63\% SR (+8\%) and 57\% SPL (+5\%) on the test unseen split\footnote{R2R Leaderboard: \href{https://evalai.cloudcv.org/web/challenges/challenge-page/97/overview}{https://evalai.cloudcv.org/web/challenges/challenge-page/97/overview}, REVERIE Leaderboard:\\ \href{https://eval.ai/web/challenges/challenge-page/606/overview}{https://eval.ai/web/challenges/challenge-page/606/overview} \label{footnote_leaderboard}}. Comparing to PRESS \cite{li2019robust} and PREVALENT \cite{hao2020towards} which only fine-tune a pre-trained BERT for extracting language features, adding recurrence into V\&L BERT and using the model directly as the navigator network allows the VLN learning to adequately benefit from the pre-trained knowledge. Such performance gain cannot be achieved by using pre-trained V\&L BERT only as a feature extractor, as will be shown in \S 4.2 \textit{Ablation Study}. Moreover, the large gain in SR with a slight increase in TL suggests that the agent is able to navigate both accurately and efficiently. Comparing to previous methods, we can see that the performance gap between the validation unseen and the test unseen splits is greatly reduced, which means our agent has a stronger generalisation ability to novel instructions and environments.

In REVERIE \cite{qi2020reverie} (Table~\ref{tab:reverie}), our \vlnbert~(init. OSCAR) generalises much better to unseen data. On the validation unseen split, the SR of navigation and object grounding has been absolutely improved by 11.13\% and 6.36\% respectively. On the test unseen split$^{\ref{footnote_leaderboard}}$, our method obtains 24.62\% SR and 19.48\% SPL for navigation, as well as 12.65\% RGS and 10.00\% RGSPL for REF, achieving a better performance than the previous best \cite{qi2020reverie} which applies SoTA navigator FAST \cite{ke2019tactical} for navigation and pointer MATTN \cite{yu2018mattnet} for object grounding. Compare our model with OSCAR initialisation to without, the navigation results on the validation splits are similar, but the navigation on the test split and the object grounding across all the data splits are largely improved. This result also suggests that it is possible to apply a BERT-based model for VLN and REF multi-task learning. Although the previous method has higher OSR, it is likely due to longer searching (long TL), the lower SR suggests that the agent does not know where to stop correctly. In Table~\ref{tab:reverie}, we also present the performance of \vlnbert~initialised from PREVALENT~\cite{hao2020towards}, which achieves the best result across almost all of the metrics in all dataset splits. It is very interesting to see that although PREVALENT is pre-trained on low-level R2R instructions~\cite{anderson2018vision} without other V\&L knowledge, it significantly boosts the navigation with high-level instructions as well as the object grounding in REVERIE. We hypothesis that the pre-trained knowledge provides the model with some structural priors, while the learning of REF is strongly influenced by navigation, especially at the early training stage where the target object is rarely observable by the agent.
% It is clear that the performance has a huge room for improvement, and we argue that learning about common sense (\eg structure of the environment) is crucial for navigation in REVERIE due to the high-level instructions. Although not investigated in this paper, we suggest that future work can pre-train \vlnbert~ for common sense knowledge to improve the performance.

\begin{figure}[t]
  \centering
  \includegraphics[width=\columnwidth]{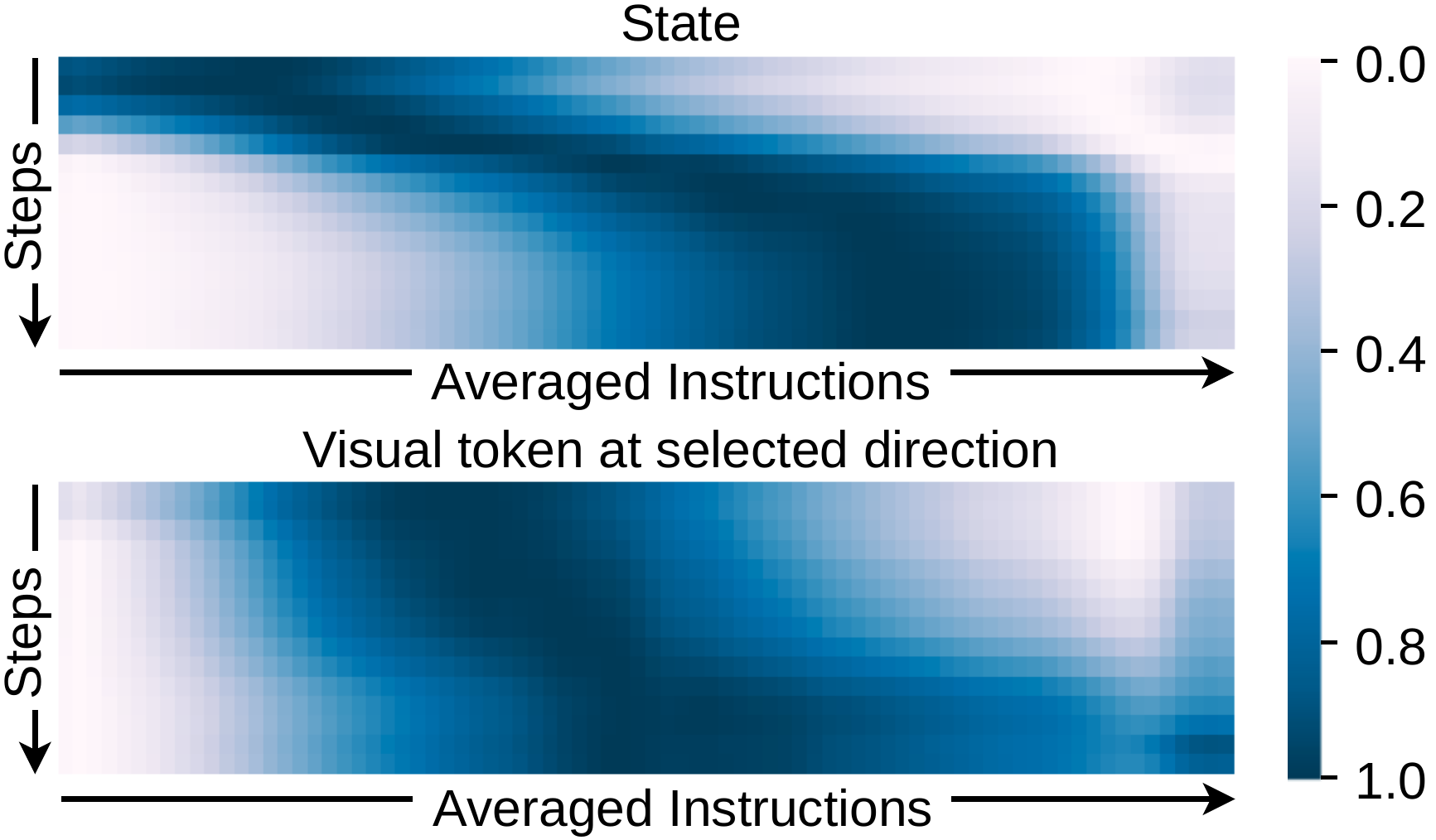}
  \caption{Averaged attention weights over all instructions in validation unseen split during navigation. \textit{State}: Attention weights with respect to the state representation. \textit{Selected Action}: Attention weights with respect to the visual token at the selected direction.}
  \label{fig:attn_weights}
  \vspace{-1.5em}
\end{figure}

\paragraph{Visualisation of Language Attention}
To demonstrate that our \vlnbert~ (init. OSCAR) is able to trace the navigation progress, we visualise the changes of language attention weights at the final Transformer layer over all instructions during navigation (Fig.~\ref{fig:attn_weights}). As the agent moves forward, the attention weights with respect to state shifts from the beginning of the instructions to the end. Since the sub-instructions and sub-paths for each sample in R2R is monotonically aligned \cite{hong-etal-2020-sub}, our results indicate that the state nicely records the partial instruction that has been completed. In terms of the attention weights with respect to the visual token at the select direction, it follows a similar pattern meaning that the most relevant part of the instruction is used for guiding the action selection.

% % -----------------------------------------------

\subsection{Ablation Study}

\begin{table*}[t]
  \begin{center}
  \resizebox{\textwidth}{!}{
  \begin{tabular}{l|cccc|c|c|rrrr|rrrr}
    \hline \hline
    \multicolumn{1}{c}{\multirow{2}{*}{Models}} & \multicolumn{6}{|c}{V\&L BERT (init. OSCAR)} & \multicolumn{4}{|c}{R2R Validation Seen} & \multicolumn{4}{|c}{R2R Validation Unseen} \Tstrut\\
    \cline{2-15} &
    \multicolumn{1}{c}{Language} & \multicolumn{1}{c}{Vision} & \multicolumn{1}{c}{State} & \multicolumn{1}{c}{Decision} & \multicolumn{1}{|c}{Matching} & \multicolumn{1}{|c}{Train} & \multicolumn{1}{|c}{TL} & \multicolumn{1}{c}{NE$\downarrow$} & \multicolumn{1}{c}{SR$\uparrow$} & \multicolumn{1}{c}{SPL$\uparrow$} & \multicolumn{1}{|c}{TL} & \multicolumn{1}{c}{NE$\downarrow$} & \multicolumn{1}{c}{SR$\uparrow$} & \multicolumn{1}{c}{SPL$\uparrow$} \Tstrut\\
    \hline \hline
    Baseline \cite{tan2019learning} & & & & & & & 11.84 & 4.44 & 57.79 & 52.85 & 12.50 & 5.26 & 49.81 & 43.45 \\
    \hline
    1 & \checkmark & & & & & & 10.81 & 4.98 & 49.95 & 46.19 & 11.34 & 5.68 & 44.15 & 39.64 \\
    2 & \checkmark & & & & & \checkmark & 11.73 & 4.18 & 59.26 & 54.12 & 12.59 & 5.00 & 52.11 & 45.75 \\
    3 & \checkmark & \checkmark & & & & & 9.26 & 6.85 & 34.77 & 33.33 & 8.92 & 7.43 & 30.74 & 29.05 \\
    4 & \checkmark & \checkmark & & & & \checkmark & 11.37 & 3.50 & 67.97 & 63.94 & 12.98 & 4.73 & 54.75 & 48.31 \\
    5 & \checkmark & \checkmark & \checkmark & & & \checkmark & 11.10 & 3.81 & 65.52 & 61.24 & 12.20 & 4.62 & 55.21 & 49.72 \\
    6 & \checkmark & \checkmark & \checkmark & \checkmark & & \checkmark & 10.70 & 3.21 & 70.32 & 66.45 & 11.46 & 4.48 & 57.22 & 52.57 \\
    \hline
    Full model & \checkmark & \checkmark & \checkmark & \checkmark & \checkmark & \checkmark & 10.79 & \textbf{3.11} & \textbf{71.11} & \textbf{67.23} & 11.86 & \textbf{4.29} & \textbf{58.71} & \textbf{53.41} \\
    \hline \hline
  \end{tabular}}
\end{center}
\caption{Ablation experiments on the effect of applying V\&L BERT for learning navigation. Checkmarks indicate using V\&L BERT to replace or to add the corresponding network component in the baseline model. \textit{Matching} indicates the \textit{cross-modal matching} (Eq.~\ref{eqn:matching}), and \textit{Train} with checkmark means the V\&L BERT is fine-tuned for navigation.}
% Network size is the number of learning parameters (M: millions), including the BERT word embeddings \cite{devlin2019bert} and the Critic in A2C \cite{mnih2016asynchronous}.}
  \label{tab:abl_arch}
\end{table*}

\begin{table}[t]
    \begin{center}
    \resizebox{\columnwidth}{!}{
    \begin{tabular}{l|cccc|cccc|c|r}
        \hline \hline
        \multicolumn{1}{c}{\multirow{2}{*}{Models}} & \multicolumn{4}{|c}{R2R Validation Seen} & \multicolumn{4}{|c|}{R2R Validation Unseen} & Batch &  \multicolumn{1}{c}{\multirow{2}{*}{Memory}} \Tstrut\\
        \cline{2-9} &
        \multicolumn{1}{c}{TL} & \multicolumn{1}{c}{NE$\downarrow$} & \multicolumn{1}{c}{SR$\uparrow$} & \multicolumn{1}{c}{SPL$\uparrow$} & \multicolumn{1}{|c}{TL} & \multicolumn{1}{c}{NE$\downarrow$} & \multicolumn{1}{c}{SR$\uparrow$} & \multicolumn{1}{c|}{SPL$\uparrow$} & Size \Tstrut\\
        \hline \hline
        Emb-Attn   & 10.84 & 3.40 & 67.19 & 63.19 & 11.31 & 4.64 & 55.60 & 51.00 & 6 & 11.0GB \\
        Init-Attn  & 10.21 & 3.93 & 61.80 & 58.06 & 10.39 & 4.59 & 53.47 & 49.05 & 6 & 11.0GB \\
        Re-Attn    &  8.99 & 5.81 & 40.84 & 39.44 &  8.85 & 6.22 & 37.76 & 35.78 & 6 & 11.0GB \\
        \hline
        Ours       & 10.79 & \textbf{3.11} & \textbf{71.11} & \textbf{67.23} & 11.86 & \textbf{4.29} & \textbf{58.71} & \textbf{53.41} & \textbf{16} & \textbf{9.2GB} \\
        \hline \hline
    \end{tabular}}
    \end{center}
    \caption{Comparison of performing language self-attention: On the raw word embeddings at each step (\textit{Emb-Attn}), on the initialised language features at each step (\textit{Init-Attn}), on the output language features from previous step (\textit{Re-Attn}), or only at initialisation (\textit{Ours}). \textit{Memory} is the training time GPU memory cost.}
    \label{tab:abl_attn}
    \vspace{-10pt}
\end{table}

\paragraph{Network Components}
Table~\ref{tab:abl_arch} shows comprehensive ablation experiments on the influence of using V\&L BERT (init. OSCAR) to replace or to add the key network components in the baseline model (EnvDrop \cite{tan2019learning}). The baseline model consists of a language encoder, a visual encoder, a state
LSTM and a decision making module, corresponding to the columns of \textit{Language}, \textit{Vision}, \textit{State} and \textit{Decision} in Table~\ref{tab:abl_arch}, respectively. E.g.,  Model \#3 replaces the language encoder and visual encoder in baseline with a V\&L BERT. For fair comparison, all models in the table are trained with the same data and training strategy as our \vlnbert. 
As the results suggested, our proposed method is a multi-functional framework, the more network components it covers, the larger performance gain it achieves. Comparing the baseline with Model \#1 and \#2, we can see that employing a pre-trained BERT as language encoder improves the performance only if the BERT is fine-tuned for navigation. This finding is also supported by using the V\&L BERT to encode both the textual and visual signals (Model \#3 and \#4). However, simply using pre-trained V\&L BERT as text and image encoders does not fully utilise its power; model \#5 indicates that relying on the original architecture of BERT to learn recurrence is feasible and it is able to achieve better results. Moreover, using the averaged visual attention weights of the final layer of the Transformer as the action probabilities (Model \#6) and enhancing the state representation with visual-textual matching as defined in Eq.~\ref{eqn:matching} (full model) further improves the agent's performance. 

\vspace{-10pt}
\paragraph{Self-Attended Language Features}
Due to the long instructions and episodes, high memory cost during training is one of the key issues that prevents previous research to apply BERT for self-attention at every time step. To demonstrate the influence of self-attending textual features during navigation, we compare the agent's performance and the training time GPU memory consumption (constrained to a single 11GB memory GPU) of re-attending the language at each step. As shown in Table~\ref{tab:abl_attn}, training for \textit{Emb-Attn}, \textit{Init-Attn} and \textit{Re-Attn} consume much more memory for each sample than performing language self-attention only at initialisation (\textit{Ours}), and their performances are worse than \textit{Ours}. 
The results of \textit{Re-Attn} degenerates significantly because at each time step the output language features aggregate the most relevant visual-textual clues at a certain viewpoint, which suppresses the valuable information in other part of the instruction for the future steps. We refer \textit{Appendix} \S C.2 for experiment on language self-attention with larger batch size by applying gradient accumulation.
% These results suggest that running self-attention on language at each step is unnecessary.

% % -----------------------------------------------

\begin{figure}[tp!]
  \centering
  \includegraphics[width=\columnwidth]{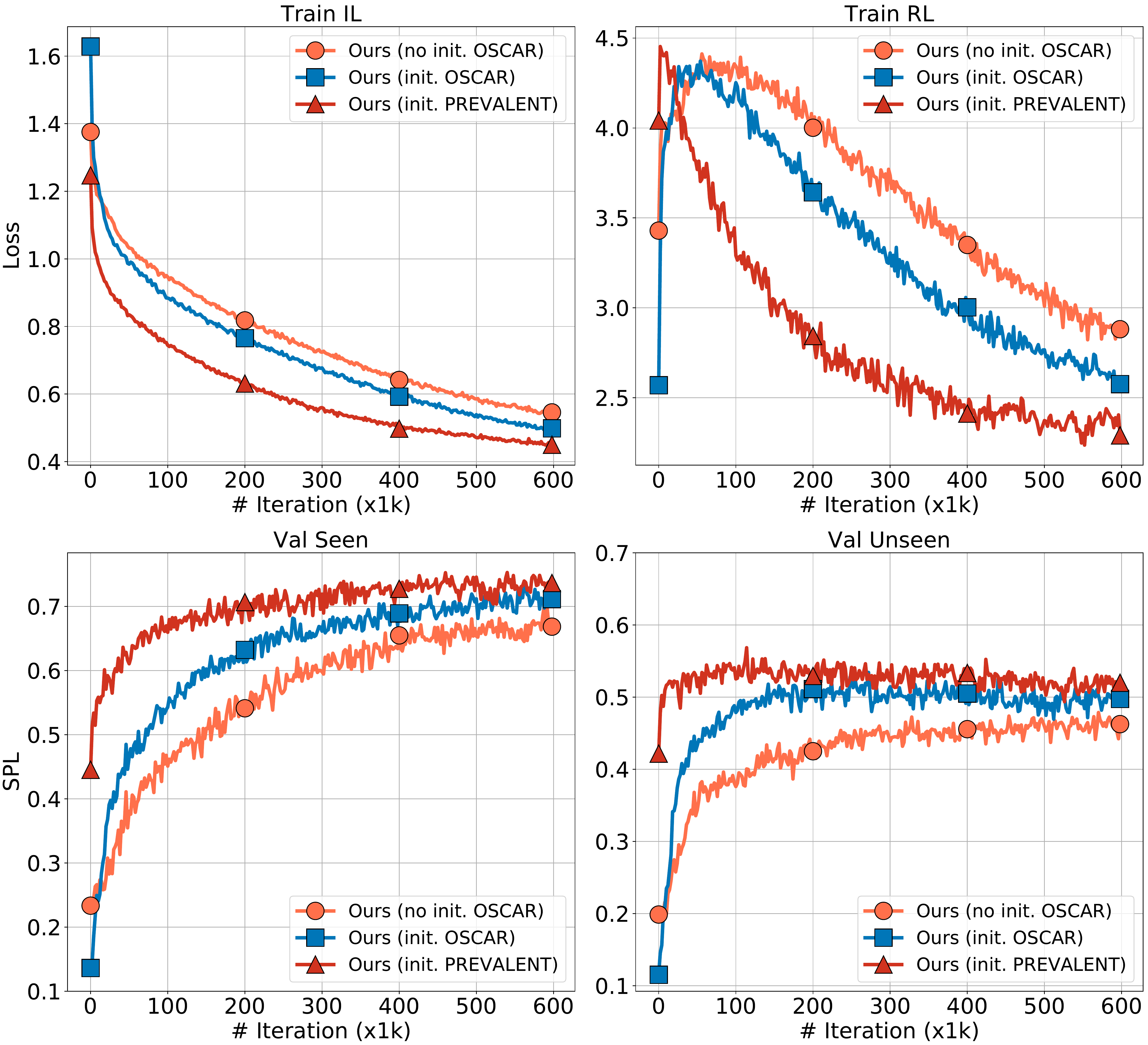}
  \caption{Comparison of the learning curves. \textit{no init.} means randomly initialised network parameters.}
  \label{fig:train_curve}
  \vspace{-10pt}
\end{figure}

\paragraph{Learning Curves}
As shown in Fig.~\ref{fig:train_curve}, we compare the learning curves of \vlnbert~ initialised from different models. The training losses of our method initialised from pre-trained OSCAR \cite{li2020oscar} converges faster than a randomly initialised model, and it reaches much higher SPL in both validation seen and unseen environments. Moreover, our model initialised from PREVALENT~\cite{hao2020towards} learns significantly faster than the other two methods and it is able to achieve a much better performance within much fewer iterations. In terms of training in real time, the model init. OSCAR takes about 7 days\footnote{Time includes evaluation on the validation splits every 2,000 iterations. Matterport 3D Simulator v0.1 is applied, whereas the latest version supports batches of agents so it is much more efficient. \label{footnote_time}} to complete 600,000 iterations of training (best result achieved in 3.5 days), while the model init. PREVALENT takes about 4.5 days$^{\ref{footnote_time}}$ (best result achieved in 1 day). Using wall-clock time in Fig.~\ref{fig:train_curve} as the x-axis will enhance the discrepancy between the three models. These results suggest that the pre-trained generic visiolinguistic knowledge is beneficial to the learning of VLN, and pre-training especially for navigation skills allows the agent to learn better in fine-tuning.

% ------------------------------------------------

\section{Conclusion} 
\label{sec:conclusion}
In this paper, we introduce recurrence into Vision-and-Language BERT and rely on its original architecture to recognise time-dependent inputs. Such innovation allows V\&L BERT to address problems with a partially observable Markov decision process, and allows the learning of downstream tasks to adequately benefit from the pre-trained generic V\&L knowledge. For VLN, our proposed \vlnbert~ applies BERT itself as the navigator network, which achieves SoTA performance in R2R \cite{anderson2018vision} and REVERIE \cite{qi2020reverie}. Moreover, results suggest that V\&L BERT with recurrence is capable of VLN and REF multi-tasking.

\vspace{-15pt}
\paragraph{Future Work} As suggested by the significant improvement on R2R \cite{anderson2018vision} achieved by our \vlnbert, we expect that the model can improve the performance of other navigation settings such as street navigation \cite{chen2019touchdown} and navigation in continuous environments \cite{krantz2020navgraph}. In this paper, we only apply our recurrent BERT for VLN. However, we believe that it has a huge potential in addressing other tasks which require sequential interactions/decisions, such as language and visual dialog \cite{byrne2019taskmaster,das2017visual, kottur2019clevr,lee2019multi,Rastogi2020TowardsSM}, dialog navigation \cite{de2018talk,nguyen2019help,nguyen2019vision} and action anticipation for reactive robot response \cite{aliakbarian2018viena,furnari2019would,koppula2015anticipating}.

{\small
\bibliographystyle{ieee_fullname}
\bibliography{egbib}
}

\clearpage

\appendix
\section*{Appendices}

\section{Datasets}

We evaluate the performance of our proposed model on two distinct datasets for VLN:
\begin{itemize}
    \item Room-to-Room (R2R) \cite{anderson2018vision}: The agent is required to navigate in photo-realistic environments (Matterport3D \cite{chang2017matterport3d}) to reach a target following low-level natural language instructions. Most of the previous works apply a panoramic action space for navigation \cite{fried2018speaker}, where the agent jumps among viewpoints pre-defined on the connectivity graph of the environment. The dataset contains 61 scenes for training; 11 and 18 scenes for validation and testing, respectively, in unseen environments.
    
    \item REVERIE \cite{qi2020reverie}: The agent needs to first navigate to a point where the target object is visible, then, it needs to identify the target object from a list of given candidates. In REVERIE, the navigational instructions are high-level while the instructions for object grounding are very specific. The dataset has in total 4,140 target objects in 489 categories, and each target viewpoint has 7 objects with 50 bounding boxes in average.
\end{itemize}

\section{Implementation Details}
We provide the implementation details of preparing visual features (\S 3.3\footnote{Link to Section 3.3 in Main Paper. \label{section}}), decision making (\S 3.3), adaptation to PREVALENT \cite{hao2020towards} (\S 3.3 \& 3.5), adaptation to REVERIE \cite{qi2020reverie} (\S 3.3 \& 3.5), critic function and reward shaping in reinforcement learning (\S 3.4).

\subsection{Visual Features (\texorpdfstring{$\boldsymbol{\S}$}~3.3$^{\ref{section}}$)}
Navigation in R2R \cite{anderson2018vision} and REVERIE \cite{qi2020reverie} are conducted in the Matterport3D Simulator \cite{chang2017matterport3d}. At each navigable viewpoint in the environment, the agent observes a $360^\circ$ panorama, consisting of 36 single-view images at 12 headings ($30^\circ$ separation) and 3 elevation angles ($\pm30^\circ$). 

\paragraph{Scene Features}
In our experiments, we only consider the scene features (grid features of the single-view images provided by ResNet-152 \cite{he2016deep} pre-trained on Places365 \cite{zhou2017places}) at the navigable directions as visual features to \vlnbert. Each visual feature is direction-aware, which is formulated by the concatenation of the convolutional image feature $\boldsymbol{f}^{v,i}_{t}\in\mathbb{R}^{2048}$ and the directional encoding $\boldsymbol{d}^{i}_{t}\in\mathbb{R}^{128}$ as:
\begin{equation}
\boldsymbol{I}^{v,i}_{t} = \left[\boldsymbol{f}^{v,i}_{t};\boldsymbol{d}^{i}_{t}\right]
\label{eqn:scene_feat}
\end{equation}
The directional encoding is formed by replicating vector $(\text{cos}\theta^{i}_{t}, \text{sin}\theta^{i}_{t}, \text{cos}\phi^{i}_{t}, \text{sin}\phi^{i}_{t})$ by 32 times, where $\theta^{i}_{t}$ and $\phi^{i}_{t}$ represents the heading and elevation angles of the image with respect to the agent's orientation \cite{fried2018speaker, tan2019learning}. Moreover, the action features $\boldsymbol{a}_{t}$ which is fed to the state representation is exactly the directional encoding at the selected direction $\boldsymbol{d}^{c}_{t}$, where $c=a^{s}_{t}$.

\paragraph{Object Features}
In REVERIE \cite{qi2020reverie}, the target objects can appear at any single-view of the panorama, we extract the object features (regional features encoded by Faster-RCNN \cite{ren2015faster} pre-trained on Visual Genome \cite{krishna2017visual} by Anderson~\etal~\cite{anderson2018bottom}) according to the positions of objects provided in REVERIE \cite{qi2020reverie}. The object features are position- and direction-aware, formulated as
\begin{equation}
\boldsymbol{I}^{o,k}_{t} = \left[\boldsymbol{f}^{o,k}_{t};\boldsymbol{p}^{k}_{t}\boldsymbol{W}^{p};\boldsymbol{d}^{k}_{t}\right]
\label{eqn:object_feat}
\end{equation}
where $\boldsymbol{f}^{o,k}_{t}\in\mathbb{R}^{2048}$ is the convolutional object features, $\boldsymbol{d}^{k}_{t}\in\mathbb{R}^{128}$ is the directional encoding of the single-view which contains the object. $\boldsymbol{p}^{k}_{t}$ represents the spatial position of the object within the image, as in MATTN \cite{yu2018mattnet}, we apply $\boldsymbol{p}^{k}_{t}=\left[\frac{x_{tl}}{W},\frac{y_{tl}}{H},\frac{x_{br}}{W},\frac{y_{br}}{H},\frac{w{\cdot}h}{W{\cdot}H}\right]$, where $(x_{tl}, y_{tl})$ and $(x_{br}, y_{br})$ are the top-left and bottom-right coordinates of the object, $(w,h)$ and $(W,H)$ are the width and height of the object and the image, respectively. The matrix $\boldsymbol{W}^{p}\in\mathbb{R}^{5\times128}$ is a learnable linear projection.

% --------------------------------------

% \subsection{Data Augmentation}
% Remove this section, argue a fair comparison with PREVALENT, and the fact that aug data comes from the train set.

\begin{figure*}[htp!]
  \centering
  \includegraphics[width=\textwidth]{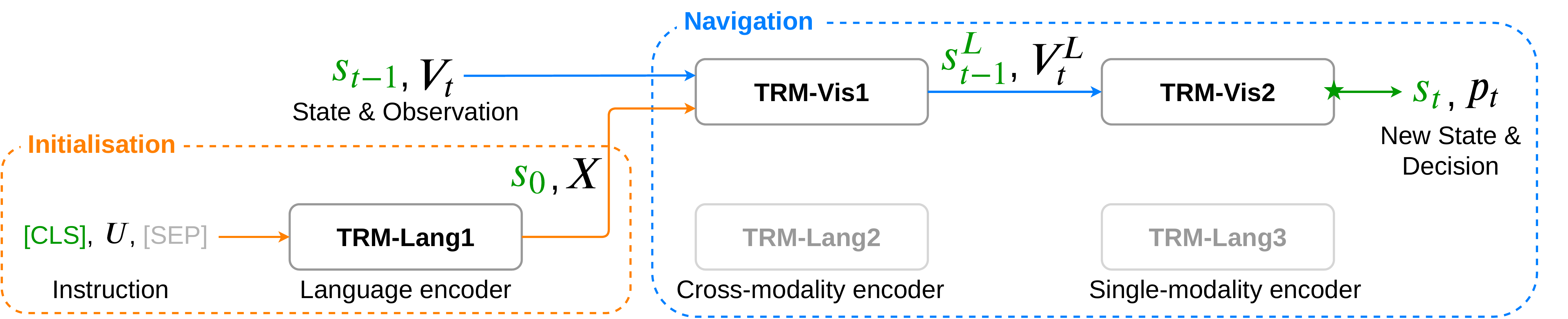}
  \caption{Adaptation to recurrent PREVALENT. At initialisation, the entire instruction is encoded by a language transformer (\textit{TRM-Lang1}), where the output feature of the \texttt{[CLS]} token servers as the initial state representation of the agent. During navigation, the concatenated sequence of state, encoded language and new visual observation are fed to the cross-modality and the single-modality encoders to obtain the updated state and decision probabilities. The updated state and the language encoding from initialisation will be fused and applied as input at the next time step. The green star (\star) indicates the \textit{cross-modal matching} and the \textit{past decision encoding} (\texorpdfstring{$\boldsymbol{\S}$}~3.3).}
  \label{fig:prevalent}
\end{figure*}

% --------------------------------------

\subsection{Decision Making (\texorpdfstring{$\boldsymbol{\S}$}~3.3)}
In R2R \cite{anderson2018vision}, there are two types of decisions that an agent infers during navigation; it either selects a navigable direction to move, or it decides to \texttt{stop} at the current viewpoint. As in most of the previous work, stopping in \vlnbert~is implemented by adding a zero vector $\boldsymbol{v}^{\text{stop}}_{t}$ to the list of visual features at navigable directions \cite{fried2018speaker,hong2020graph,ma2019self,tan2019learning}, as
\begin{equation}
\boldsymbol{V}_{\!t} = \langle{\boldsymbol{v}^{1}_{t},\boldsymbol{v}^{2}_{t},...,\boldsymbol{v}^{n}_{t}, \boldsymbol{v}^{\text{stop}}_{t} \mid \boldsymbol{v}^{i}_{t} \in \mathbb{R}^{2176}}\rangle
\label{eqn:stop_feat}
\end{equation}
Our \vlnbert~determines to stop by predicting the largest attention score to the \texttt{stop} representation at the final transformer layer. Otherwise, the agent will move to a navigable direction with the largest score.

However, in REVERIE \cite{qi2020reverie}, we directly apply the attention scores over the candidate objects for stopping. To be specific, the visual tokens in REVERIE consists of sequences of scene features and object features:
\begin{equation}
\langle{\boldsymbol{v}^{1}_{t},\boldsymbol{v}^{2}_{t},...,\boldsymbol{v}^{n}_{t},\boldsymbol{o}^{1}_{t},\boldsymbol{o}^{2}_{t},...,\boldsymbol{o}^{m}_{t} \mid \boldsymbol{v}^{i}_{t}\in\boldsymbol{V}_{\!t},\boldsymbol{o}^{k}_{t}\in\boldsymbol{O}_{\!t}}\rangle
\label{eqn:revis_feat}
\end{equation}
When the model predicts larger attention scores for at least one of the object token than all of the scene tokens, the agent will stop and will select the object with the largest score as the grounded object for REF. Such formulation has two advantages; First, it relates the object searching process to navigation, \ie, the agent should not stop navigation if it has low confidence of localising the target object. Second, it allows the reinforcement learning to benefit the object grounding, since the action logits for \texttt{stop} is the greatest attention scores over objects.

% --------------------------------------

\subsection{Adaptation to PREVALENT (\texorpdfstring{$\boldsymbol{\S}$}~3.3 \& \texorpdfstring{$\boldsymbol{\S}$}~3.5)}

As shown in Fig.~\ref{fig:prevalent}, we adapt our \vlnbert~to the LXMERT-like \cite{tan2019lxmert} architecture in PREVALENT \cite{hao2020towards}. At initialisation, the transformer \textit{TRM-Lang1} encodes the instruction $\boldsymbol{U}$ and uses the output features of the \texttt{[CLS]} token to represent agent's initial state. During navigation, the concatenated sequence of the previous state $\boldsymbol{s}_{t-1}$, the encoded language from initialisation $\boldsymbol{X}$ and the new visual observation $\boldsymbol{V}_{\!t}$ will be fed to the \textit{cross-modality encoder} to obtain the language-aware state feature $\boldsymbol{s}^{L}_{t-1}$ and the language-aware visual features $\boldsymbol{V}^{L}_{\!t}$. Finally, \textit{TRM-Vis2} will process $\boldsymbol{s}^{L}_{t-1}$ and $\boldsymbol{V}^{L}_{\!t}$ to produce a new state $\boldsymbol{s}_{t}$ and a decision $\boldsymbol{p}_{t}$. 

Pre-training of PREVALENT \cite{hao2020towards} applies the outputs from \textit{TRM-Lang3} for attended masked language modelling and action prediction, but fine-tuning on R2R \cite{anderson2018vision} only use the output language features from \textit{TRM-Lang1} and rely on a downstream network EnvDrop \cite{tan2019learning} for navigation. In contrast, our method does not require any downstream network, we leverage the visual transformers \textit{TRM-Vis1} and \textit{TRM-Vis2} to learn state-language-vision relationship for better decision marking.

\paragraph{Cross-Modal Matching}
The \textit{cross-modal matching} for \textit{State Refinement} (see Eq.~12 in \S 3.3) is applied to enhance the state representation of PREVALENT-based \vlnbert. Since the final transformer \textit{TRM-Vis2} only process the state and visual features, we apply the averaged attention scores over the visual features with respect to $\boldsymbol{s}_{t}$ to weight the input visual tokens $\boldsymbol{V}_{t}$, but apply the averaged attention scores over the textual features with respect to $\boldsymbol{s}^{L}_{t-1}$ to weight the input language tokens $\boldsymbol{X}$. Results in Table \ref{tab:pre_match} show that cross-modal matching for state improves the agent's performance in unseen environments.

\begin{table}[hbt!]
    \begin{center}
    \resizebox{\columnwidth}{!}{
    \begin{tabular}{l|cccc|cccc}
        \hline \hline
        \multicolumn{1}{c}{\multirow{2}{*}{Models}} & \multicolumn{4}{|c}{R2R Validation Seen} & \multicolumn{4}{|c}{R2R Validation Unseen} \Tstrut\\
        \cline{2-9} &
        \multicolumn{1}{c}{TL} & \multicolumn{1}{c}{NE$\downarrow$} & \multicolumn{1}{c}{SR$\uparrow$} & \multicolumn{1}{c}{SPL$\uparrow$} & \multicolumn{1}{|c}{TL} & \multicolumn{1}{c}{NE$\downarrow$} & \multicolumn{1}{c}{SR$\uparrow$} & \multicolumn{1}{c}{SPL$\uparrow$} \Tstrut\\
        \hline \hline
        w/o Matching   & 10.87 & \textbf{2.44} & \textbf{76.79} & \textbf{73.13} & 11.72 & 4.08 & 62.07 & 56.15 \\
        with Matching  & 11.13 & 2.90 & 72.18 & 67.72 & 12.01 & \textbf{3.93} & \textbf{62.75} & \textbf{56.84} \\
        \hline \hline
    \end{tabular}}
    \end{center}
    \caption{Performance of PREVALENT-based \vlnbert~with and without cross-modal matching for agent's state.}
    \label{tab:pre_match}
\end{table}

% --------------------------------------

\subsection{Critic Function (\texorpdfstring{$\boldsymbol{\S}$}~3.4)}
We apply the A2C \cite{mnih2016asynchronous} for reinforcement learning. At each time step, the critic, a multi-layer perceptron, predicts an expected value from the updated state representation as:
\begin{equation}
e_{t}=(\text{ReLU}(\boldsymbol{s}_{t}\boldsymbol{W}^{E1}))\boldsymbol{W}^{E2}
\label{eqn:critic}
\end{equation}
where ReLU is the Rectified Linear Unit function, $\boldsymbol{W}^{E1}$ and $\boldsymbol{W}^{E2}$ are learnable linear projections.

% --------------------------------------

\subsection{Reward Shaping (\texorpdfstring{$\boldsymbol{\S}$}~3.4)}

In addition to the progress rewards defined in EnvDrop \cite{tan2019learning}, we apply the normalised dynamic time warping \cite{ilharco2019general} as a part of the reward to encourage the agent to follow the instruction to navigate. Moreover, we introduce a negative reward to penalise the agent if it misses the target.

\paragraph{Progress Reward}
As formulated in the EnvDrop \cite{tan2019learning}, we apply the progress rewards as strong supervision signals for directing the agent to approach the target. To be specific, let $D_{t}$ to be the distance from agent to target at time step $t$, and ${\Delta}D_{t}=D_{t}-D_{t-1}$ to be the change of distance by action $a_{t}$, the reward at each step ($a_{t}{\neq}\texttt{stop}$) is defined as:
\begin{equation}
r^{D,step}_{t}=
\begin{cases}
    +1.0,     & {\Delta}D_{t}>0.0 \\
    -1.0,     & \text{otherwise}
\end{cases}
\label{eqn:reward_step}
\end{equation}
When the agent decides to stop ($a_{t}{=}\texttt{stop}$), a final reward is assigned depending on if the agent successfully complete the task:
\begin{equation}
r^{D,final}_{t}=
\begin{cases}
    +2.0,     & D_{t}<3.0 \\
    -2.0,     & \text{otherwise}
\end{cases}
\label{eqn:reward_stepf}
\end{equation}
Overall, the agent will receive a positive reward if it approaches the target and completes the task (by stopping within 3 meters to the target viewpoint), while it will be penalised with a negative reward if it moves away from the target or stops at a wrong viewpoint.

\paragraph{Path Fidelity Rewards}
The \textit{Progress Reward} encourages the agent to approach the target, but it does not constrain the agent to take the shortest path. As there could be multiple routes to the target, the agent could learn to take a longer path or even a cyclic path to maximise the total reward. To address the problem, we apply the normalised dynamic time warping reward \cite{ilharco2019general}, a measurement of the similarity between the ground-truth path and the predicted path, to urge the agent to follow the instruction accurately. Let $P_{t}$ be the normalised dynamic time warping \cite{ilharco2019general} at time $t$, and ${\Delta}P_{t}=P_{t}-P_{t-1}$ to be the change of $P$ caused by action $a_{t}$, the reward for $a_{t}{\neq}\texttt{stop}$ is defined as:
\begin{equation}
r^{P,step}_{t}={\Delta}P_{t}
\label{eqn:reward_ndtw}
\end{equation}
and the reward for $a_{t}{=}\texttt{stop}$:
\begin{equation}
r^{P,final}_{t}=
\begin{cases}
    +2.0P_{t},     & D_{t}<3.0 \\
    +0.0,          & \text{otherwise}
\end{cases}
\label{eqn:reward_ndtwf}
\end{equation}

Moreover, as suggested by many previous works, there exists a large discrepancy between the agent's oracle success rate and success rate, indicating that it does not learn well to stop accurately. To address this issue, we introduce a negative stopping reward $r^{S}_{t}$ which will be triggered whenever the agent first approaches the target but then departs from it. To be precise, if $D_{t-1}{\leq}1.0$ and ${\Delta}D_{t}>0.0$:
\begin{equation}
r^{S}_{t}=-2.0{\times}(1.0-D_{t-1})
\label{eqn:reward_stop}
\end{equation}

The normalised dynamic time warping reward $r^{P}_{t}$ and the stopping reward $r^{S}_{t}$ together form the path fidelity rewards which can encourage the agent to navigate efficiently.
In summary, the overall reward at each step during navigation can be expressed as:
\begin{equation}
r_{t}=
\begin{cases}
    r^{D,step}_{t}+r^{P,step}_{t}+r^{S}_{t},  & a_{t}{\neq}\texttt{stop} \\
    r^{D,final}_{t}+r^{P,final}_{t},          & a_{t}{=}\texttt{stop}
\end{cases}
\label{eqn:reward_all}
\end{equation}

\paragraph{Ablation Study}

We perform ablation experiments on training the OSCAR-based and the PREVALENT-based \vlnbert~with and without the path fidelity rewards. As shown in Table~\ref{tab:abl_rewards}, the models trained with the path fidelity rewards achieve higher Success Rate (SR) and lower Trajectory Length (TL), leading to higher Success weighted by Path Length (SPL). Despite the improvements in SR, the gap between the Oracle Success Rate (OSR) and SR is kept roughly the same for the PREVALENT-based model and is reduced by about 1.45\% the OSCAR-based models. These results suggest that the path-fidelity rewards benefit the agent to navigate more accurate and efficient. Note that, comparing to the results that are shown in Table 1 and Table 3 of our Main Paper, such reward shaping only contributes to a slight gain of the improvement, whereas the structure of the recurrent BERT is much more influential.

\begin{table}[htb!]
    \begin{center}
    \resizebox{\columnwidth}{!}{
    \begin{tabular}{l|cccc|cccc}
        \hline \hline
        \multicolumn{1}{c}{\multirow{2}{*}{Models}} & \multicolumn{4}{|c}{R2R Validation Seen} & \multicolumn{4}{|c}{R2R Validation Unseen} \Tstrut\\
        \cline{2-9} &
        \multicolumn{1}{c}{TL} & \multicolumn{1}{c}{OSR} & \multicolumn{1}{c}{SR$\uparrow$} & \multicolumn{1}{c}{SPL$\uparrow$} & \multicolumn{1}{|c}{TL} & \multicolumn{1}{c}{OSR} & \multicolumn{1}{c}{SR$\uparrow$} & \multicolumn{1}{c}{SPL$\uparrow$} \Tstrut\\
        \hline \hline
        OSCAR &  &  &  &  &  &  &  &  \\
        \quad $r^{D}$ only & 11.15 & 77.86 & 70.71 & 66.64 & 12.62 & 66.92 & 58.32 & 52.48 \\
        \quad $r^{D}+r^{P}+r^{S}$ & 10.79 & 76.79 & \textbf{71.11} & \textbf{67.23} & 11.86 & 65.86 & \textbf{58.71} & \textbf{53.41} \\
        \hline
        PREVALENT &  &  &  &  &  &  &  &   \\
        \quad $r^{D}$ only & 12.24 & 77.28 & 69.93 & 64.46 & 12.89 & 69.52 & 62.15 & 55.65 \\
        \quad $r^{D}+r^{P}+r^{S}$ & 11.13 & 78.16 & \textbf{72.18} & \textbf{67.72} & 12.01 & 70.24 & \textbf{62.75} & \textbf{56.84} \\     
        \hline \hline
    \end{tabular}}
    \end{center}
    \caption{Performance of OSCAR-based and PREVALENT-based \vlnbert~trained with and without the path fidelity rewards.}
    \label{tab:abl_rewards}
\end{table}

\section{Language Attention}

\subsection{Additional Explanation of the Averaged Language Attention Weights (\texorpdfstring{$\boldsymbol{\S}$}~4.1)}

In Figure 3 of the Main Paper, we visualised the averaged attention weights over all instructions in validation unseen split during navigation. Notice that in this visualisation, we interpolate the instruction to 80 words and the trajectory to 15 steps for each sample to compute the averaged attention weights. Since a large portion of instructions in R2R are less than 80 words, the last few tokens in the plot are corresponding to the full stop punctuation and the [SEP] token, which are less likely to provide valuable information to the state and hence receive very little weight. However, for the last step in the bottom plot, the selected visual token for stopping could have a high correspondence to the full stop and [SEP] which are strong indications of the end of navigation.

\subsection{Self-Attended Language Features with Gradient Accumulation (\texorpdfstring{$\boldsymbol{\S}$}~4.2)}

To evaluate the influence of performing different language self-attention under the same batch size, we train \textit{Emb-Attn}, \textit{Init-Attn} and \textit{Re-Attn} with gradient accumulation to achieve batch size of 16 (same as \textit{Ours}) while keeping learning rate unchanged. Comparing to the results reported in Table 4 of the Main Paper, the performance of all the three methods has been significantly improved. On the validation unseen split, \textit{Emb-Attn} even outperforms \textit{Ours} which does not re-attend the language at each time step. However, a downside for accumulating gradient is that the training speed is significantly reduced, which is about 2.5 times slower for accumulating 4 batches of size 4. Nevertheless, this result suggests that agent's performance can be further improved by training with larger batch size, potentially including \textit{Ours}. Therefore, whether it is necessary to perform language self-attention at each navigational step when there is sufficient computational power remains a question worth investigating. We will leave it as a future work.

\begin{table}[htb!]
    \begin{center}
    \resizebox{\columnwidth}{!}{
    \begin{tabular}{l|cccc|cccc|c|r}
        \hline \hline
        \multicolumn{1}{c}{\multirow{2}{*}{Models}} & \multicolumn{4}{|c}{R2R Validation Seen} & \multicolumn{4}{|c|}{R2R Validation Unseen} & Accu- &  \multicolumn{1}{c}{\multirow{2}{*}{Speed}} \Tstrut\\
        \cline{2-9} &
        \multicolumn{1}{c}{TL} & \multicolumn{1}{c}{NE$\downarrow$} & \multicolumn{1}{c}{SR$\uparrow$} & \multicolumn{1}{c}{SPL$\uparrow$} & \multicolumn{1}{|c}{TL} & \multicolumn{1}{c}{NE$\downarrow$} & \multicolumn{1}{c}{SR$\uparrow$} & \multicolumn{1}{c|}{SPL$\uparrow$} & Grad \Tstrut\\
        \hline \hline
        Emb-Attn   & 11.36 & \textbf{2.97} & 70.91 & 66.26 & 12.22 & \textbf{4.15} & \textbf{61.00} & \textbf{54.85} & \checkmark & 1.43 \\
        Init-Attn  & 12.11 & 3.78 & 64.84 & 59.49 & 12.32 & 4.37 & 58.62 & 52.53 & \checkmark & 1.33 \\
        Re-Attn    &  9.71 & 4.44 & 54.16 & 51.51 & 10.19 & 5.28 & 47.89 & 44.64 & \checkmark & 1.38 \\
        \hline
        Ours       & 10.79 & 3.11 & \textbf{71.11} & \textbf{67.23} & 11.86 & 4.29 & 58.71 & 53.41 & \xmark & \textbf{3.40} \\
        \hline \hline
    \end{tabular}}
    \end{center}
    \caption{Comparison of performing language self-attention with gradient accumulation. We set the batch size of the first three methods as 4 and accumulate the gradient for 4 iterations, so that the overall batch size for each update is 16, same as \textit{Outs}. Unit for \textit{Speed} is 1,000 iterations per hour (including the time spent on evaluation).}
    \label{tab:abl_attn_grad_accu}
\end{table}

% --------------------------------------

\section{Visualisation (\texorpdfstring{$\boldsymbol{\S}$}~4.1)}

\begin{figure*}[t]
  \centering
  \includegraphics[width=0.92\textwidth]{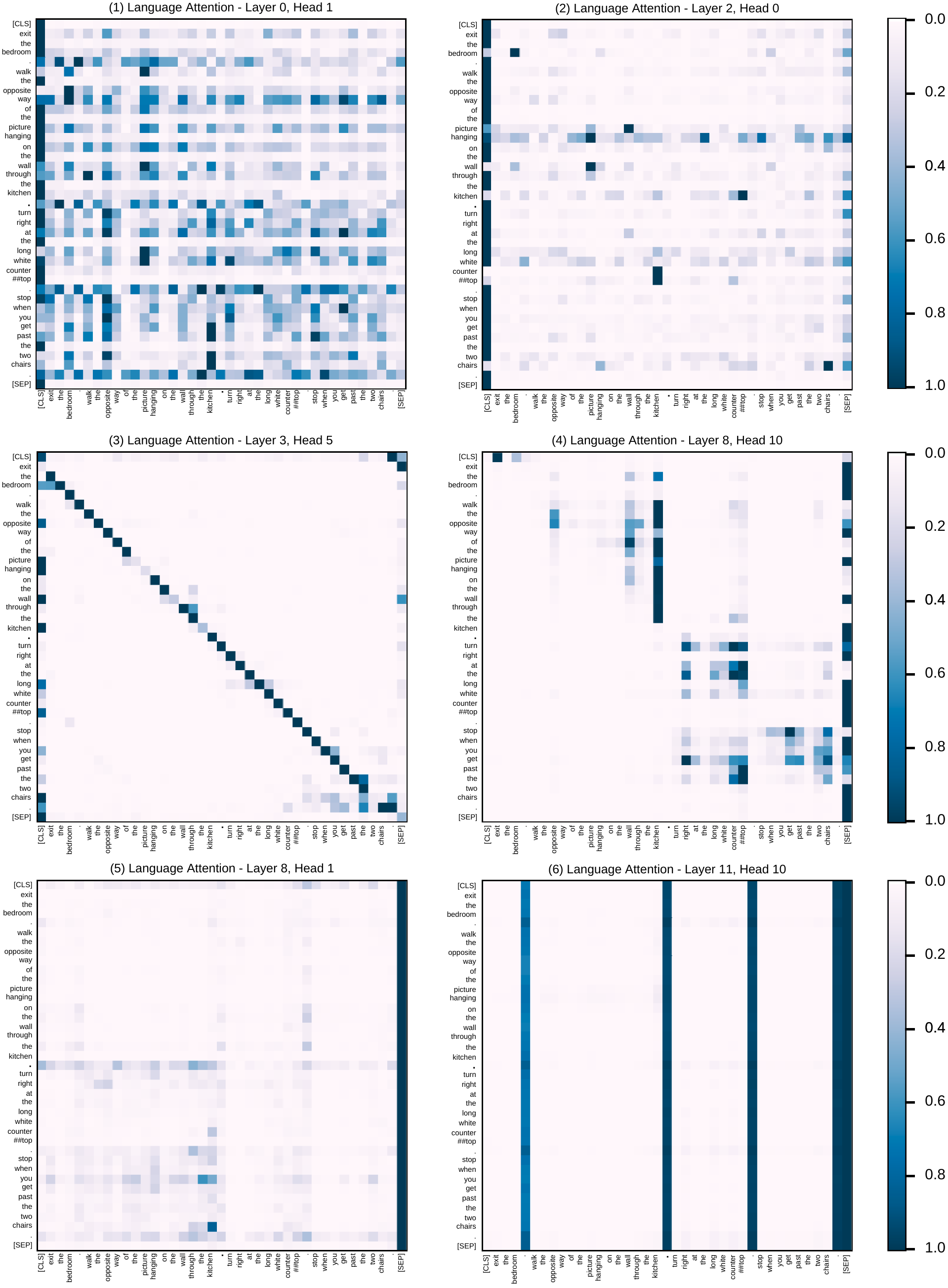}
  \caption{Language self-attention weights of some selected heads at initialisation. The attention weights are normalised for each row.}
  \label{fig:supp_f1}
\end{figure*}

\begin{figure*}[t]
  \centering
  \includegraphics[width=\textwidth]{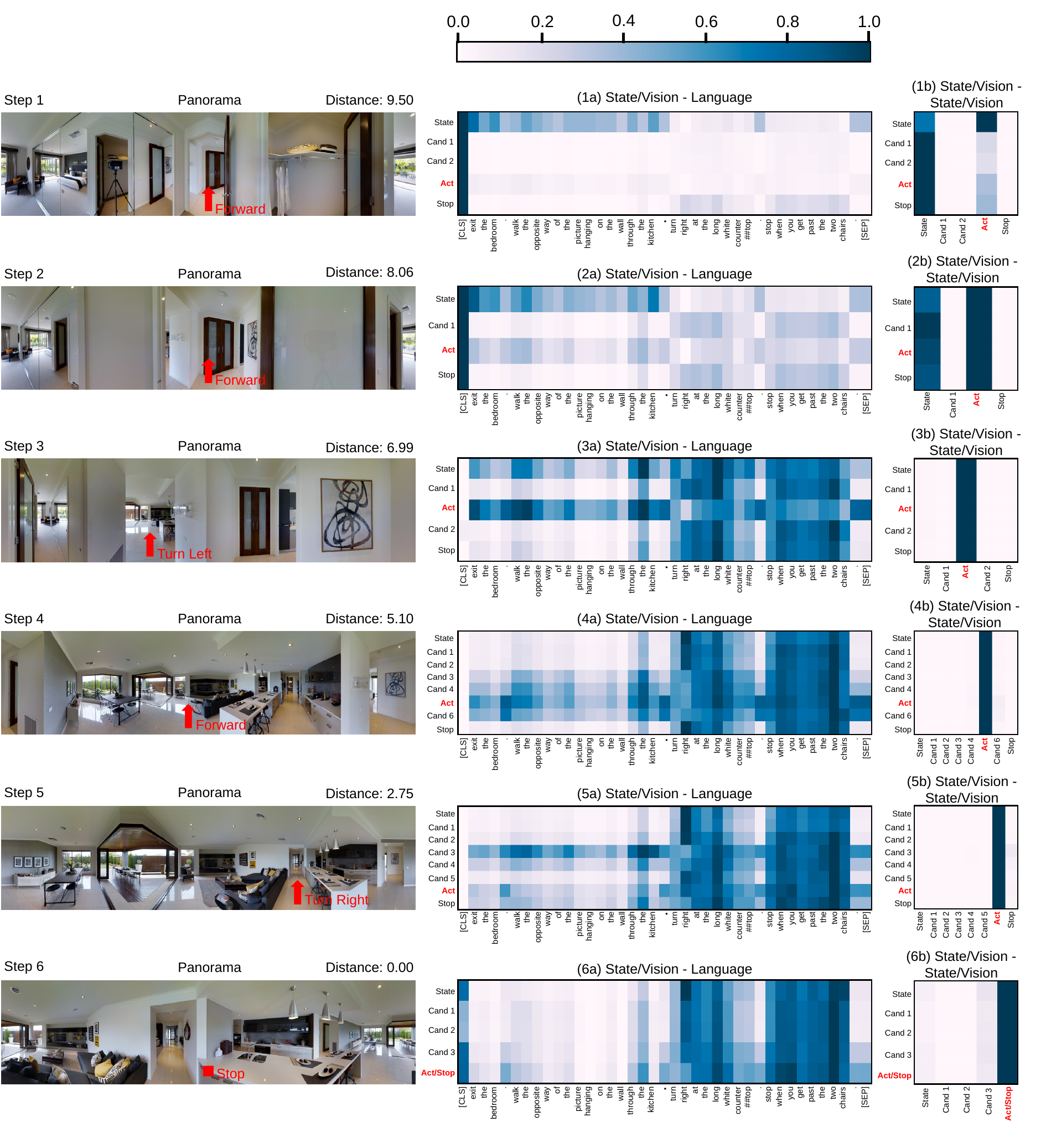}
  \caption{Visualisation of a trajectory and the averaged attention weights at the final ($l{=}12$-th) layer. The centre of each panorama is roughly the agent's heading direction at the corresponding time step. \textit{Distance} is the agent's distance to target in meters. The attention weights are normalised for each row. \textcolor{red}{Texts in red} indicates the predicted action (corresponds to a candidate view) at each time step.}
  \label{fig:supp_f2}
\end{figure*}

\begin{figure*}[t]
  \centering
  \includegraphics[width=\textwidth]{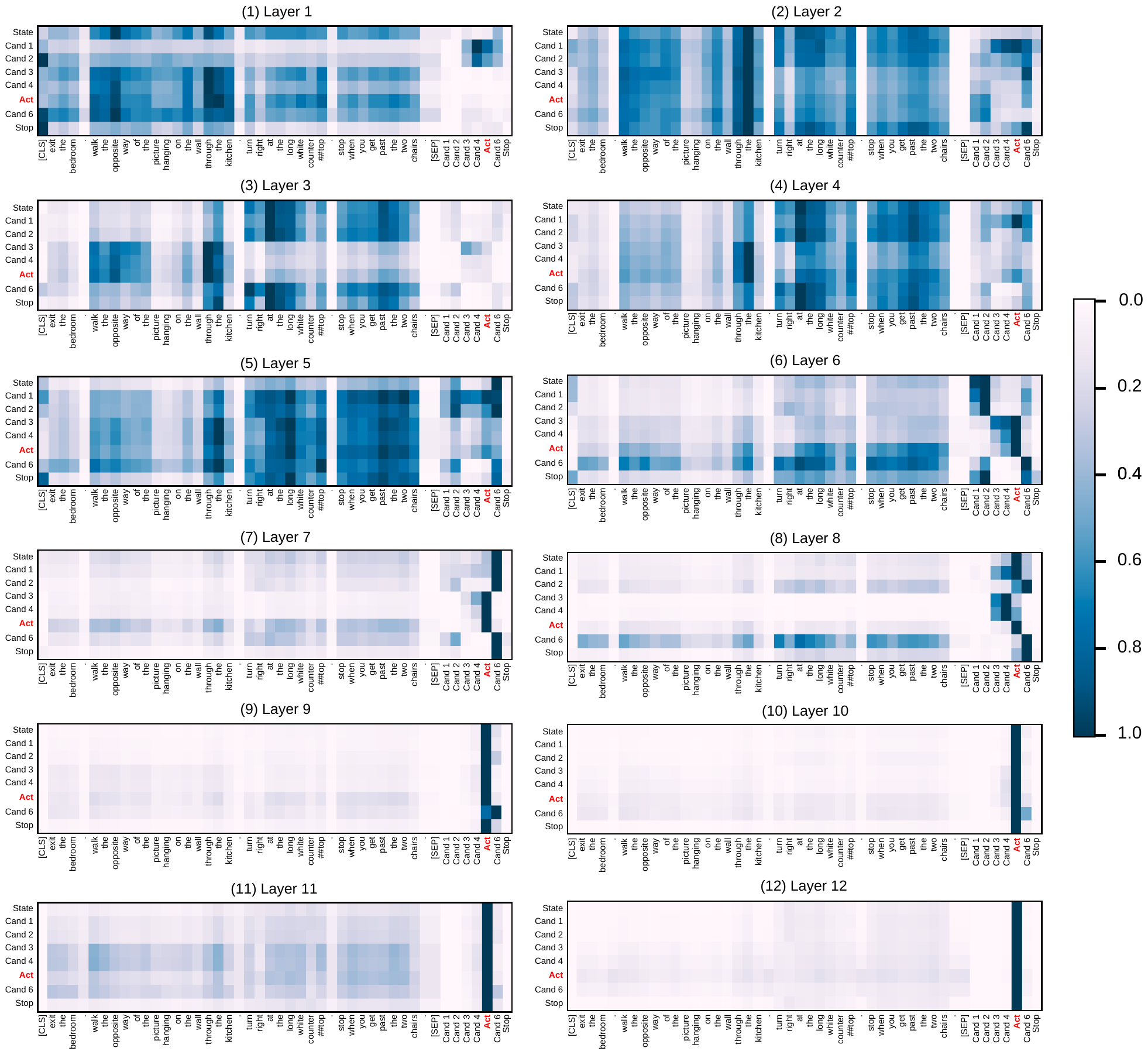}
  \caption{Averaged state/vision-to-state/vision/language attention weights at each layer at Step 4 of the trajectory. The attention weights are normalised for each row. \textcolor{red}{Texts in red} indicates the predicted action (corresponds to a candidate view) at the current time step ($t{=}4$).}
  \label{fig:supp_f3}
\end{figure*}

As shown in Figure~\ref{fig:supp_f1}, \ref{fig:supp_f2} and \ref{fig:supp_f3}, we visualise the language-to-language and the state/vision-to-state/vision/language attention weights of a sample in the validation unseen split. As shown by the panoramas in Fig.~\ref{fig:supp_f2} and the given instruction ``\textit{Exit the bedroom. Walk the opposite way of the picture hanging on the wall through the kitchen. Turn right at the long white countertop. Stop when you get past the two chairs.}'', the agent needs to understand the complex contextual clues, including different scene clues (\textit{bedroom}, \textit{kitchen}), different object clues (\textit{picture}, \textit{wall}, \textit{countertop}, \textit{chair}) and various directional clues (\textit{forward}, \textit{opposite}, \textit{left/right}, \textit{stop}) to complete the task.

\paragraph{Language Self-Attention}
Fig.~\ref{fig:supp_f1} shows the language self-attention attention weights at some selected heads at initialisation ($t{=}0$); different heads demonstrate different functions and the attentions at different layers behave very differently. Plot (1) and (4) show the general pattern of attentions at shallow (Layer 0) and deep layers (Layer 8); words in shallow layers tend to collect information from the entire sentence, while words at deep layers have higher correspondence to the adjacent words since they are semantically more relevant. It is very interesting to see that the attention head in Plot (2) learns to attend the adjectives and the action-related terms, which describe the objects and scenes, for the initialised state representation (\texttt{[CLS]}). This head also learns about the co-occurrence between different entities, for example, \textit{picture} has higher correspondence to \textit{wall}, and \textit{countertop} has higher correspondence to \textit{kitchen}. In contrast, the attention head in Plot (3) learns to extract the important landmarks such as \textit{bedroom}, \textit{picture}, \textit{kitchen} and \textit{chairs}. Attention head in Plot (5) learns about the \texttt{[SEP]} token, which indicates the ending of the instruction. Attention weights in Plot (6) show the most frequent pattern of the attention heads at the final ($l{=}12$-th) layer, those heads seem to aggregate information from the punctuations in the instruction. This implies the heads could have learnt about breaking the sentence into multiple sub-sentences. Refer to the idea of sub-instruction proposed by Hong~\etal~\cite{hong-etal-2020-sub}, such attention pattern could be beneficial for matching the current observation to a particular and the most relevant part of the instruction.

\paragraph{State/Vision Step-wise Attention}
Fig.~\ref{fig:supp_f2} shows the trajectory of the agent starting from a bedroom, taking a series of actions and eventually stopping at the target location. It also displays the averaged attention weights at the final ($l{=}12$-th) layer for the state and visual tokens. Note that the averaged attention for state/vision tokens is representative since we apply the averaged attention weights for \textit{visual-textual matching} to state and use it as the action probabilities (see Eq.~12 and \textit{Decision Making} in \S 3.3, respectively). As shown in Plot (1a-6a), the attention shifts from the beginning of the instruction to the end, which agrees with the agent's navigation progress. It is also interesting to see that at the final layer, the state token is more influential to the predicted action at the first two steps, while the language tokens are more influential at the later steps. The state/vision self-attention in Plot (1b-6b) reflects the action prediction at each time step. The first row in each plot (attention of state with respect to candidate actions) shows the prediction result, we can see that the agent is very confident in each decisions. Moreover, starting from Step 3, the information from state and from different views are aggregated to support choosing the correct direction.

\paragraph{State/Vision Layer-wise Attention}
To better understand how the visual and language features are aggregated to support action prediction, we visualise the layer-wise attention at Step 4 of the trajectory ($t{=}4$). As shown by Fig.~\ref{fig:supp_f3} Plot (1-12), at the first two layers, candidate views collect information from the entire instruction. But as the signals propagate to deeper layers (Layer 3-6), the visual tokens attend more the middle part of the instruction, which should be more relevant to the current observations. Interestingly, starting from Layer 6, the visual features tend to dominate the attention, and information aggregates toward the visual token at the predicted direction (which is the correct direction). We can see that, at Layer 7-9, the network still has some doubts about the candidate directions that are spatially closer to the correct direction. But after implicitly reasoning in deeper layers, the network becomes very confident about choosing the correct direction.

\clearpage

\clearpage

\clearpage

\end{document}